\documentclass[final,3p,times]{elsarticle}

\usepackage{geometry}

\usepackage{amssymb}
\usepackage{amsmath,bm}
\usepackage{amsthm}
\usepackage{amsmath}
\usepackage{relsize}
\usepackage{bbm}
\usepackage[linesnumbered, ruled, lined]{algorithm2e}
\usepackage{multicol}
\usepackage{graphicx}
\usepackage{booktabs}
\usepackage{subfigure}
\usepackage{latexsym}
\usepackage{textcomp}
\biboptions{numbers,sort&compress}

\begin{document}

\begin{frontmatter}

\title{Deep clustering with fusion autoencoder\tnoteref{label1}}



\author[add1]{Shuai Chang\corref{cor1}}
\ead{aerocs@my.swjtu.edu.cn}
\cortext[cor1]{Corresponding author}

\address[add1]{School of Computing and Artificial Intelligence, Southwest Jiaotong University, Chengdu, 611756, China}
%

\begin{abstract}
Embracing the deep learning techniques for representation learning in clustering research has attracted broad attention in recent years, yielding a newly developed clustering paradigm, viz. the deep clustering (DC). Typically, the DC models capitalize on autoencoders to learn the intrinsic features which facilitate the clustering process in consequence. Nowadays, a generative model named variational autoencoder (VAE) has got wide acceptance in DC studies. Nevertheless, the plain VAE is insufficient to perceive the comprehensive latent features, leading to the deteriorated clustering performance. In this paper, a novel DC method is proposed to address this issue. Specifically, the generative adversarial network and VAE are coalesced into a new autoencoder called fusion autoencoder (FAE) for discerning more discriminative representation that benefits the downstream clustering task. Besides, the FAE is implemented with the deep residual network architecture which further enhances the representation learning ability. Finally, the latent space of the FAE is transformed to an embedding space shaped by a deep dense neural network for pulling away different clusters from each other and collapsing data points within individual clusters. Experiments conducted on several image datasets demonstrate the effectiveness of the proposed DC model against the baseline methods.
\noindent 
\end{abstract}


\begin{highlights}
\item Two generative frameworks, i.e., variational autoencoder and generative adversarial network are seamlessly united to construct a new autoencoder: the fusion autoencoder.
\item Residual neural network is adopted to build the network architecture of the proposed autoencoder.
\item Clusters are captured by a separate dense neural network.
\end{highlights}


\end{frontmatter}

\section{Introduction} 

In the unsupervised learning domain, clustering is a fundamental data analysis task which partitions a group of objects in such a way that, according to some principles, similar objects are assigned to a partition while different partitions contain dissimilar objects. Clustering has extensive real-world applications, e.g., detecting communities among the social media networks \cite{INUWADUTSE202164}, segmenting images via clustering the semantic features \cite{JIAO202083}, and helping enterprises discovering the social audience clusters in a business information network \cite{ZHENG2020107126}. Clustering algorithms have continuously emerged in the past few decades. For instance, k-means \cite{macqueen1967some} and spectral clustering \cite{10.5555/2980539.2980649} are two well-known classical clustering methods that have been widely used in  data processing \cite{BIENVENIDOHUERTAS2021102829, GUO2021107166, 6909417, 6889674}. Nevertheless, traditional clustering models are either barely able to uncover semantic similarities or have huge time complexity when dealing with high-dimensional, large-scale, and high-semantic data, making it unfeasible to apply these algorithms on such sophisticated datasets \cite{MRABAH2020206}. To overcome these difficulties, dimensionality reduction strategies such as principal component analysis and matrix factorization, non-linear transformation techniques e.g. kernel method, are integrated into clustering models for extracting essential features of the objects, which improve the efficiency and performance consequently \cite{JAFARZADEGAN20191, LI2012120, CHANG2021106807}. In spite of that, these methods are either prone to discriminative feature loss or have limited abilities to fit the complex real-world data \cite{9099972}.

Deep learning has prevailed among artificial intelligence research in recent years thanks to its exceptional representation learning capability. For instance, deep neural networks can be treated as universal function approximators that approximating any measurable function to an arbitrary degree of accuracy \cite{HORNIK1989359}. Clustering methods that take advantage of the deep learning techniques for jointly learning hidden features of the data are referred to as the deep clustering (DC) models \cite{XU2020106260}. Typically, DC algorithms can be classified into three categories: (1) direct cluster optimization; (2) autoencoder based models; (3) generative model based methods. The direct cluster optimization DC models only focus on the clustering loss to optimize the deep neural network, such as \cite{8237888, 7780925, 8590804} leverage convolutional neural networks to uncover the representation and exploit affinities between data points for guiding the clustering procedure. However, clustering without the decoder may cause the training process to be corrupted \cite{9099972}. On the other hand, the DC frameworks presented in \cite{ijcai2017-243, Dizaji_2017_ICCV, MORADIFARD2020185, YU2020216} harness autoencoders that simultaneously optimize clustering and reconstruction losses for preserving the local structures, and therefore preventing the embedded space from being distorted \cite{ijcai2017-243}. The main drawback of utilizing autoencoders in DC algorithms is that the latent spaces of autoencoders are apt to be compact, which could trigger the clusters to overlap with each other \cite{8622629}. 

While the generative model based DC algorithms utilize the generative models to carry out the clustering process. Nowadays, a probabilistic graph based generative model called variational autoencoder (VAE) \cite{DBLP:journals/corr/KingmaW13} has drawn many researchers' attention \cite{JOO2020107514, HOU2019183, takahashi2019variational, NEURIPS2019_7d12b66d}. Practical applications of VAE include fault diagnosis \cite{YAN2021107142, ARIASCHAO2021324}, real-world grade trajectory generation \cite{CHEN2021332}, and extracting latent features for disease classification \cite{GUNDUZ2021102452}. Technically, VAE compresses the input data to a constrained latent distribution space and then maximizes the lower bound on the likelihood of the data to minimize the reconstruction loss. Although VAE can learn data distributions in the latent space which is good for unsupervised learning tasks, the Gaussian prior may lead to crowding clusters, hindering the subsequent clustering process to effectively separate different groups \cite{8957256}. Several DC models \cite{10.5555/3172077.3172161, 9010011, 9207523, 9207493} turn to the Gaussian mixture prior for modeling the discrete clusters in the latent space, which have contributed to a more clustering-oriented VAE framework. In spite of that, vanilla VAE lacks the ability that captures the detailed features of objects and is notorious for blurry output images \cite{10.5555/2969033.2969125, pmlr-v48-larsen16, DBLP:conf/iclr/DumoulinBPLAMC17}, which could potentially lead to inferior clustering performance. Another generative model which is called generative adversarial network (GAN) empirically learns the map that transforms the latent variables to the complex data distribution by playing a min-max game. Some recently developed DC approaches \cite{8954410, mukherjee2019clustergan} exploit the GAN framework for data clustering. Apart from this, extensive researches have demonstrated that GAN can produce high-quality images \cite{DBLP:conf/iclr/BrockDS19} and perceive the diversities and consistencies lying in the latent space when coupled with an encoder framework \cite{10.5555/3305890.3305928, DBLP:conf/iclr/DumoulinBPLAMC17}. Motivated by these advantages, we assume that with the assistance of GAN, VAE could mitigate the blurry image issue and learn the comprehensive latent features which are helpful for enhancing the clustering performance.

Based on these considerations, in this paper, a novel DC method called Deep Clustering with Fusion AutoEncoder (DCFAE) is proposed to overcome the aforementioned challenges when performing clustering with VAE. Firstly, we decide to use the mean vectors for the clustering task instead of changing the prior distribution of the latent variables. The reasons behind this strategy are twofold: (1) we observe that the mean vectors of the Gaussian latent variables are dispersed among relatively separate clusters; (2) the mean vectors carry the definite latent information of the original datapoints, suggesting that the mean vectors can properly represent the corresponding input data. Secondly, we merge VAE with GAN to form a new autoencoder, namely the Fusion AutoEncoder (FAE). More concretely, the original inputs and reconstructed outputs of the VAE are treated as the true and fake samples which should be correctly identified by a discriminator, that is to say, the decoder of the VAE casts the generator of the GAN at the same time. In this way, we suppose that FAE can capture more subtle discriminative latent features which are beneficial for clustering. Thirdly, FAE leverages the residual neural network architecture \cite{7780459} which allows FAE to be built with deeper neural networks for extracting the high-level features and therefore improves the representation learning ability even further. And finally, a fully connected neural network is implemented to shift the latent space of the mean vector to a more k-means clustering friendly embedding space, which amplifies the inter-cluster separation and condenses the intra-cluster points by minimizing the cross-entropy between two distributors measured by pairwise similarities among the data points inside the latent mean vector space and the embedding space, respectively.

The main contributions of this study are listed as follows:

\begin{itemize}
  \item Two generative models VAE and GAN are directly bonded together to forge the FAE for improving the performance of VAE and mining more detailed latent features that benefit the clustering downstream task.
  \item The network architecture of FAE is built with deep residual convolutional networks which enable FAE to capture deeper latent features and refine the representation learning ability consequently.
  \item The mean vector latent space of the FAE is converted to a k-means clustering friendly embedding space by a deep dense neural network. Experiments conducted on several image datasets show the effectiveness of the designed DCFAE clustering method and confirm the contributions of the proposed novelties to the clustering performance.
\end{itemize}

The rest of this paper are originated as follows. Section 2 presents the related work of this study. The proposed clustering method is detailed in Section 3. Section 4 demonstrates a series of experiments and we conclude the paper in Section 5.

\section{Related work}

In this section, we first review the fundamentals about VAE and GAN together with several merging strategies of these two frameworks and then introduce some DC algorithms.

\subsection{Variational autoencoder and generative adversarial network}

Variational AutoEncoder (VAE) can be viewed as two independently parametrized models: the recognition model and the generative model, a.k.a the encoder and the decoder. Let $\boldsymbol{x}$ and $\boldsymbol{z}$ denote the input data and the latent variable, respectively. The encoder aims to stochastically approximate the true posterior distribution of the latent variable through an inference model $q_{\boldsymbol{\phi}} (\boldsymbol{z} \vert \boldsymbol{x})$ parameterized by $\boldsymbol{\phi}$. Conversely, the decoder tries to learn a distribution $p_{\boldsymbol{\theta}}(\boldsymbol{x} \vert \boldsymbol{z})$ parameterized by $\boldsymbol{\theta}$ that generates datapoints by sampling latent variables from the prior distribution $p(\boldsymbol{z})$. The marginal likelihood can be expressed as:

\begin{equation}\label{Eq-177-1}
\begin{split}
\mathop{log} p_{\boldsymbol{\theta}} (\boldsymbol{x}) & = \mathbb{E}_{q_{\boldsymbol{\phi}} (\boldsymbol{z} \vert \boldsymbol{x})} \mathop{log} p_{\boldsymbol{\theta}} (\boldsymbol{x}) \\
& \geq \mathbb{E}_{q_{\boldsymbol{\phi}} (\boldsymbol{z} \vert \boldsymbol{x})} \left( \mathop{log} p_{\boldsymbol{\theta}} (\boldsymbol{x}\ \vert\ \boldsymbol{z}) - \mathop{log} \frac{q_{\boldsymbol{\phi}} (\boldsymbol{z}\ \vert\ \boldsymbol{x})}{p(\boldsymbol{z})} \right) = \mathcal{L}_{\boldsymbol{\phi}, \boldsymbol{\theta}} (\boldsymbol{x}).
\end{split}
\end{equation}

The term in the second line of Eq. \eqref{Eq-177-1} is commonly known as the Evidence Lower Bound (ELBO). When both the encoder and decoder are implemented using deep neural networks, ELBO can be optimized by Stochastic Gradient Descent (SGD) with the reparameterization trick \cite{DBLP:journals/corr/KingmaW13}.

The Generative Adversarial Network (GAN) also contains two separate models, viz. the generator and the discriminator. The generator transforms the noise variables $\boldsymbol{z}$ to the data space by the mapping $G_{\boldsymbol{\nu}}(\boldsymbol{z})$ parametrized by $\boldsymbol{\nu}$. On the other hand, the discriminator $D_{\boldsymbol{\psi}}(\boldsymbol{x})$ plays a role in identifying whether an input comes from the true data. The generator and the discriminator are constructed by deep neural networks and they are concurrently trained in such a manner that, the generator makes an effort to fool the discriminator by generating more realistic outputs while the discriminator progressively becomes better at telling the true and fake samples apart. The training procedure can be formulated as the following mini-max game:

\begin{equation}\label{Eq-188-2}
\mathop{min}_{G_{\boldsymbol{\nu}}} \mathop{max}_{D_{\boldsymbol{\psi}}} L(G_{\boldsymbol{\nu}}, D_{\boldsymbol{\psi}}) = \mathbb{E}_{p_{\mathcal{D}}(\boldsymbol{x})} \mathop{log} D_{\boldsymbol{\psi}}(\boldsymbol{x}) + \mathbb{E}_{p(\boldsymbol{z})} \left( \mathop{log} \left( 1 - D_{\boldsymbol{\psi}}\left(G_{\boldsymbol{\nu}}(\boldsymbol{z})\right) \right) \right)
\end{equation}

\noindent where $p_{\mathcal{D}}(\boldsymbol{x})$ and $p(\boldsymbol{z})$ stand for the data distribution and the prior noise distribution, respectively. After been trained sufficiently, a well-designed GAN is supposed to reach an equilibrium that the discriminator cannot distinguish between the true and fake samples anymore.

Although VAE and GAN have wide adoption in recent deep learning studies, there are typical issues associated with them. For instance, compared with GAN, images generated by VAE are usually blurry, while GAN lacks the encoder-decoder architecture \cite{DBLP:conf/iclr/DumoulinBPLAMC17} and is hard to train \cite{10.1007/978-3-030-01264-9_23}. Plenty of researches focus on tackle these challenges by combining the two frameworks. \citet{HOU2019183} aim to improve the performance of VAE by using deep-wise feature reconstruction loss for capturing perceptual features and deploying GAN training mechanism to keep the generated images on the manifold of the natural images. The proposed model can generate high-quality face images with clear facial elements and the trained VAE performs well in facial attribute recognition. A similar strategy is adopted in \cite{pmlr-v48-larsen16}, except that the feature extracting is carried out by the discriminator and the decoder is also updated by the GAN loss. The work presented in \cite{DBLP:conf/iclr/DumoulinBPLAMC17} embeds inference mechanism into the GAN framework. More concretely, the discriminator is required to distinguish between two joint distributions. One is the joint distribution $q(\boldsymbol{x}, \boldsymbol{z})$ defined over the data and encoder output, the other is the joint distribution $p(\boldsymbol{x}, \boldsymbol{z})$ of the latent prior and the decoder output. The motivation is that by matching these two joint distributions, all marginals and conditional distributions shall be matched, therefore ensuring the encoder conditional distribution will match the posterior $p(\boldsymbol{z}\ \vert\ \boldsymbol{x})$. \citet{10.5555/3305890.3305928} improved the VAE by providing an arbitrarily flexible inference model with it. Specifically, instead of pre-defining the inference model $q(\boldsymbol{z}\ \vert\ \boldsymbol{x})$ as a Gaussian distribution, the posterior is implemented through a deep neural network, and a discriminator is responsible for identifying whether the pairs $(\boldsymbol{x}, \boldsymbol{z})$ come from the distribution $p_{\mathcal{D}}(\boldsymbol{x})p(\boldsymbol{z})$ or are sampled using $p_{\mathcal{D}}(\boldsymbol{x})q(\boldsymbol{z}\ \vert\ \boldsymbol{x})$. It is shown that under ideal conditions, the proposed model could learn exact maximum-likelihood parameter assignments for the generative model and obtain the exact posterior distribution for the latent variables. Despite the successes of these proposed methods, they either consist of complicated architectures or suffer from difficult training processes. In this paper, we try to merge VAE and GAN in a simple, direct, and compact way that learns the discriminative information suitable for clustering. 

\subsection{Deep clustering}

Deep clustering (DC) is referred to jointly learning representation from large-scale and high-dimensional data using the deep neural networks for the clustering task. Many DC models directly use the clustering loss for guiding the deep neural networks to learn representations. For example, \citet{8237888} employ the deep convolutional network to learn the constrained pseudo-labels of the images and then calculates the cosine distance between pseudo-labels to measure the similarities. Clustering is cast as a binary pairwise-classification framework that adaptively generates the cluster indicators through a threshold function defined on the cosine distance. The imposed constraint encourages the pseudo-labels to approach one-hot vector thus the clustering results could be obtained directly. The extension work \cite{8590804} generalizes the clustering constraint so that the indicator features can be ideally shaped  to one-hot vectors. \citet{7780925} performed the clustering and representation learning in a recurrent process. In each period, for the forward pass,  agglomerative clustering is executed to update the cluster is based on the cluster affinity; for the backward pass, representation parameters are optimized guided by the current clustering result. The local structure is considered by capturing the difference between the nearest neighbor and other neighbors of a specific cluster. Discrete representation learning methods demonstrated in \cite{10.5555/3305381.3305542} focuses on fetching the invariant representations by reducing the gap between the predictions of the augmented and original data. The clustering progress is accomplished by a deep neural network that maximizes the theoretic dependency between the inputs and outputs. 

DC models built on autoencoders generally utilize the encoder for latent feature learning and equip the reconstruction loss for preventing the feature space distortion. \citet{8622629} leverages a deep fully connected neural network for mapping the latent space of the autoencoder to an embedding space appropriate to k-means clustering. \citet{xie2016unsupervised} designed Deep Embedded Clustering (DEC) which uses k-means to initialize the cluster centers according to the non-linear deep mapping established by a stacked autoencoder and assigns embedded points to the cluster centers based on their similarities calculated by the $t$-distribution. The deep mapping and the cluster centroids are refined through minimizing the KL divergence between the soft assignment distribution and a target distribution that captures the high confidence assignments. Noted that DEC dispenses with the decoder during the clustering phase, \citet{ijcai2017-243} incorporate the reconstruction loss into the DEC framework for cluster center refinement while preserving the local structure of data. It has been demonstrated that the reconstruction loss can also stabilize the clustering process in terms of the prior knowledge \cite{YU2020216}. For sustaining the learned representations to be locally invariant, \citet{DIALLO202196} take on the contractive autoencoder and self-augmentation technique for developing a DC model suited for document datasets. The clustering progress is comparable to DEC except that a similarity measurement is adopted for maintaining the cluster region. The convolutional autoencoder based DC method \cite{Dizaji_2017_ICCV} attaches a softmax layer to the encoder for predicting the probabilistic cluster assignments which are regularized by the uniform prior. Reconstruction loss is used as a regularization term to avoid the model getting stuck in local minima and the clustering loss is calculated by the Kullback-Leibler (KL) divergence between two distributions obtained from two different encoders with shared network parameters.

Generative models such as VAE and GAN can also be utilized in DC frameworks. \citet{10.5555/3172077.3172161} replace the single Gaussian prior of the VAE with mixture of Gaussians to form a generative model (VaDE) which is apt for clustering. However, the variational posterior of the inference model is simply approximated by mean-field distribution which could decrease the clustering accuracy under specific circumstances. The VAE based DC model proposed in \cite{9207523} jointly updates the prior and posterior parameters to omit the pre-training phase which is necessary for VaDE. The Gaussian mixture model is also adopted as the prior of the encoder in \cite{8957256}, except that the parameters are learned by variational inference. \citet{Yang_2019_ICCV} combine graph embedding with VAE to incorporate local data structures, furnishing the network with powerful representation learning abilities as the global and local structural constraints are simultaneously considered. Another VAE-based DC algorithm \cite{XU2020106260} endeavors to learn useful representations and distill structure information from the observed data by maximizing the mutual information between them. The generative model GAN also has applications in DC algorithm developments. \citet{8954410} implement three networks, a generator, a discriminator, and a clusterer that play an adversarial game to synthesize real samples and learn the embedding space for clustering. Cosine similarities between the clusterer outputs are polarized to 1 (for similar outputs) or 0 (for dissimilar outputs) by a self-paced learning algorithm served as the clustering objective. Noticing that the latent points show no observable clusters in the latent space of GAN, \citet{mukherjee2019clustergan} choose a mixture of discrete and continuous prior for good interpolation and good clustering. Besides, an encoder network that can precisely recover the latent vector is employed for enhancing the clustering performance. Although these DC algorithms have got satisfactory clustering results, the motivation of this paper is uniting VAE and GAN to create a novel autoencoder for representation learning and then shifting the latent space of the autoencoder to an embedding space suitable for k-means based on a strategy similar to  presented in \cite{8622629}.

\section{The proposed method}

This section describes how the proposed DCFAE model is constructed followed by depicting its network architecture and the optimization process.

\subsection{Model formulation}

\subsubsection{The FAE model}

We first demonstrate the methodology that builds FAE. Let $\boldsymbol{x}$ denote an observed datapoint that follows the true underlying distribution $p_{\mathcal{D}}(\boldsymbol{x})$ and $\boldsymbol{z}$ stand for the latent variables, respectively. FAE generally consists of three components, namely encoder, decoder, and discriminator. The encoder maps the input observation $\boldsymbol{x}$ to the parameters that specify the posterior distribution $q_{\boldsymbol{\phi}} (\boldsymbol{z}\ \vert\ \boldsymbol{x})$. Here, $q_{\boldsymbol{\phi}} (\boldsymbol{z}\ \vert\ \boldsymbol{x})$ is treated as a factorized Gaussian distribution and thus the outputs of the encoder are the mean and variance parameters of the diagonal Gaussian. The decoder converts latent samples $\boldsymbol{z}$ to the parameters that prescribe the conditional distribution $p_{\boldsymbol{\theta}} (\boldsymbol{x}\ \vert\ \boldsymbol{z})$. Besides, the decoder is employed as the generator that adversarially competes with the discriminator. We model $p_{\boldsymbol{\theta}} (\boldsymbol{x}\ \vert\ \boldsymbol{z})$ as Bernoulli distribution and assume the latent prior $p(\boldsymbol{z})$ obeys the unit Gaussian distribution. On the other hand, the discriminator $D_{\boldsymbol{\psi}}$ seeks to spot out the genuine images from the fakes. All of these three modules are implemented by deep residual convolutional neural networks with corresponding learnable parameters $\boldsymbol{\phi}$, $\boldsymbol{\theta}$, and $\boldsymbol{\psi}$. Hence, the inference and the generative process can be expressed as:

\begin{equation}\label{Eq-208-3}
\begin{gathered}
(\boldsymbol{\mu}, \mathop{log} \boldsymbol{\sigma}) = EncoderResNet_{\boldsymbol{\phi}} (\boldsymbol{x}) \\
q_{\boldsymbol{\phi}} (\boldsymbol{z}\ \vert\ \boldsymbol{x}) = \mathcal{N} (\boldsymbol{z};\ \boldsymbol{\mu}, \mathop{diag}(\boldsymbol{\sigma})) \\
p(\boldsymbol{z}) = \mathcal{N} (\boldsymbol{z};\ \boldsymbol{0}, \boldsymbol{I}) \\
\boldsymbol{\eta} = DecoderResNet_{\boldsymbol{\theta}} (\boldsymbol{z}) \\
p_{\boldsymbol{\theta}} (\boldsymbol{x}\ \vert\ \boldsymbol{z}) = Bernoulli (\boldsymbol{x}; \boldsymbol{\eta}).
\end{gathered}
\end{equation}

Note that the log-variance output of the encoder in Eq. \eqref{Eq-208-3} aims for numerical stability. By plugging these distributions into Eq. \eqref{Eq-177-1}, the ELBO can be derived as:

\begin{equation}\label{Eq-220-4}
\begin{split}
\mathcal{L}_{ELBO} (\boldsymbol{x}) &= \mathbb{E}_{q_{\boldsymbol{\phi}} (\boldsymbol{z} \vert \boldsymbol{x})} \mathop{log} p_{\boldsymbol{\theta}} (\boldsymbol{x}\ \vert\ \boldsymbol{z}) - \mathcal{D}_{KL} \left(q_{\boldsymbol{\phi}} (\boldsymbol{z}\ \vert\ \boldsymbol{x})\ \Vert\ p(\boldsymbol{z}) \right) \\
&= \mathbb{E}_{q_{\boldsymbol{\phi}} (\boldsymbol{z} \vert \boldsymbol{x})} \mathop{log} Bernoulli (\boldsymbol{x}; \boldsymbol{\eta}) - \mathcal{D}_{KL} \left( \mathcal{N} (\boldsymbol{z}; \boldsymbol{\mu}, \mathop{diag}(\boldsymbol{\sigma}))\ \Vert\ \mathcal{N} (\boldsymbol{z}; \boldsymbol{0}, \boldsymbol{I}) \right) \\
&= \frac{1}{M} \sum\nolimits_{i = 1}^M \left[ \left( \sum\nolimits_{j = 1}^H x_j^{(i)} \mathop{log} \eta_j^{(i)} + \left(1 - x_j^{(i)}\right) \mathop{log} \left(1 - \eta_j^{(i)}\right) \right) \right.
\\
&\left. \quad - \frac{1}{2} \left( \mathop{log} \frac{1}{\mathop{det} \left( \mathop{diag} (\boldsymbol{\sigma}^{(i)}) \right)} - L + \mathop{Tr} \left (\mathop{diag} \left (\boldsymbol{\sigma^{(i)}}\right )\right ) + {\boldsymbol{\mu}^{(i)}}^T \boldsymbol{\mu}^{(i)} \right) \right] \\
&= \frac{1}{M} \sum\nolimits_{i = 1}^M \left[ \left( \sum\nolimits_{j = 1}^H x_j^{(i)} \mathop{log} \eta_j^{(i)} + \left(1 - x_j^{(i)}\right) \mathop{log} \left(1 - \eta_j^{(i)}\right) \right) - \frac{1}{2} \sum\nolimits_{k = 1}^L \left(\sigma_k^{(i)} + {\mu_k^{(i)}}^2 - \mathop{log} \sigma_k^{(i)} -1 \right) \right].
\end{split}
\end{equation}

In Eq. \eqref{Eq-220-4}, $\mathcal{D}_{KL} (\cdot \Vert \cdot)$ stands for the KL divergence between two distributions, $\mathop{Tr} (\cdot)$ denotes the matrix trace, and $\mathop{det}(\cdot)$ symbolizes the determinant of a square matrix. $M$ is the batch size, $H$ and $L$ correspond to the dimensions of $\boldsymbol{x}$ and $\boldsymbol{z}$, respectively. $x_j^{(i)}$ is the $j$-th element of $\boldsymbol{x}^{(i)}$, the same notation is applied for $\eta_j^{(i)}$. Since $\boldsymbol{z}$ is sampled from the distribution $q(\boldsymbol{z}\ \vert\ \boldsymbol{x})$, the stochastic trait of $\boldsymbol{z}$ will block out gradients backpropagating. Thanks to the reparameterization trick \cite{DBLP:journals/corr/KingmaW13}, we can circumvent the obstruction by maintaining the randomness of $\boldsymbol{z}$ through another random variable $\boldsymbol{\varepsilon}$ which follows the standard normal distribution:

\begin{equation}\label{Eq-232-5}
\begin{gathered}
\boldsymbol{\varepsilon} \sim \mathcal{N}(\boldsymbol{0}, \boldsymbol{I}) \\
\boldsymbol{z} = \boldsymbol{\mu} + \boldsymbol{\sigma} \odot \boldsymbol{\varepsilon}
\end{gathered}
\end{equation}

\noindent where $\odot$ represents the element-wise multiplication. After applying Eq. \eqref{Eq-232-5}, the ELBO $\mathcal{L}_{\boldsymbol{\phi}, \boldsymbol{\theta}}$ can be maximized through SGD.

As for the adversarial process, $0$ and $1$ are regarded as the ground truth label for the fake and real images. In each mini-batch training, we combine the generated and the real images in tandem to create a double-batch sized images $[\tilde{\boldsymbol{x}}, \boldsymbol{x}]$, accordingly, a double-batch sized label $\tau$ whose elements are arranged in $M$ $0$s ahead and $M$ $1$s behind is concatenated. Then we forward pass $[\tilde{\boldsymbol{x}}, \boldsymbol{x}]$ through the discriminator $D$ to get the predicted label $\tilde{\tau}$. As the discriminator attempts to maximize the probability that correctly identifies a given image as real or fake, we train the discriminator by minimizing the binary cross-entropy between $\boldsymbol{\tau}$ and $\tilde{\boldsymbol{\tau}}$:

\begin{equation}\label{Eq-243-6}
\mathcal{L}_D = - \frac{1}{2M} \sum\nolimits_{i = 1}^{2M} \tau_i \mathop{log} \tilde{\tau}_i + (1 - \tau_i) \mathop{log} (1 - \tilde{\tau}_i).
\end{equation}

While the generator (also the decoder of the FAE, these two terms will be interchangeably used hereafter, provided that no ambiguity exists) tries to fool the discriminator by minimizing $\mathop{log} (1 - D(G(\boldsymbol{z})))$. Instead of performing the minimization process, we adopt an alternative method stated in \cite{10.5555/2969033.2969125} that maximizes $\mathop{log} D(G(\boldsymbol{z}))$. First, the generated images are fed into the discriminator to get the predictions $\boldsymbol{\tau}'$. Then we minimize the binary cross-entropy between $\mathbbm{1}_M$ and $\boldsymbol{\tau}'$ that indirectly maximizes $\mathop{log} D(G(\boldsymbol{z}))$:

\begin{equation}\label{Eq-250-7}
\mathcal{L}_G = - \frac{1}{M} \sum\nolimits_{i = 1}^M \mathop{log} {\tau'}_i
\end{equation}

\noindent where $\mathbbm{1}_M$ is an $M$-dimensional vector whose elements are all $1$.

Based on the details expressed above, the objective function of FAE can be formulated as:

\begin{subequations}
\begin{align}
&\mathop{min}_{\boldsymbol{\phi}, \boldsymbol{\theta}} \mathcal{L}_{\alpha} = - \mathcal{L}_{ELBO} + \lambda\ \mathcal{L}_G = - \mathbb{E}_{q_{\boldsymbol{\phi}} (\boldsymbol{z}\ \vert\ \boldsymbol{x})} p_{\boldsymbol{\theta}} (\boldsymbol{x}\ \vert\ \boldsymbol{z}) + \mathcal{D}_{KL} \left( q_{\boldsymbol{\phi}} (\boldsymbol{z}\ \vert\ \boldsymbol{x})\ \Vert\ p(\boldsymbol{z}) \right) - \lambda\ \mathbb{E}_{q_{\boldsymbol{\phi}} (\boldsymbol{z}\ \vert\ \boldsymbol{x})} \mathop{log} D_{\boldsymbol{\psi}} (p_{\boldsymbol{\theta}} (\boldsymbol{x}\ \vert\ \boldsymbol{z})), \label{Eq-262-8a} \\
&\mathop{min}_{\boldsymbol{\psi}} \mathcal{L}_{\beta} = \mathcal{L}_D = - \mathbb{E}_{p_{\mathcal{D}} (\boldsymbol{x})} \mathop{log} D_{\boldsymbol{\psi}} (\boldsymbol{x}) - \mathbb{E}_{q_{\boldsymbol{\phi}} (\boldsymbol{z}\ \vert\ \boldsymbol{x})} \mathop{log} D_{\boldsymbol{\psi}} (p_{\boldsymbol{\theta}} (\boldsymbol{x}\ \vert\ \boldsymbol{z})) \label{Eq-262-8b}
\end{align}
\end{subequations}

\noindent where $\mathcal{L}_\alpha$ and $\mathcal{L}_\beta$ stand for two separate objectives which are minimized in sequence during training. $\lambda$ is a hyperparameter that balances the VAE loss and the generative adversarial loss.

\subsubsection{Deep clustering with FAE}

\begin{figure*}[!ht]
\centering
\subfigure[$\boldsymbol{\mu}$ latent space]{
\begin{minipage}[t]{0.31\textwidth}
\includegraphics[height=5.0cm,width=5.0cm]{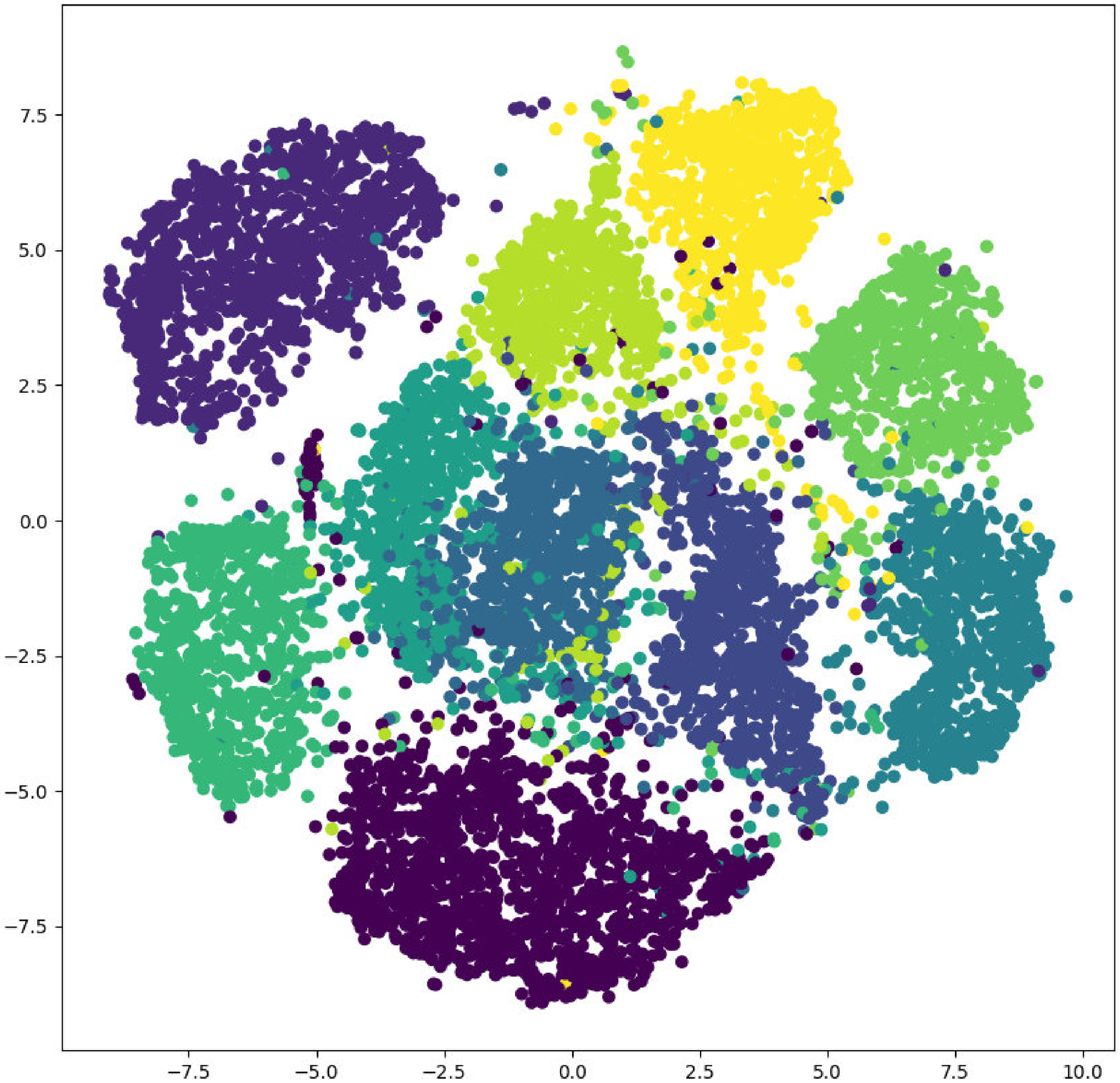}
\end{minipage}
}
\subfigure[$\boldsymbol{z}$ latent space]{
\begin{minipage}[t]{0.31\linewidth}
\includegraphics[height=5.0cm,width=5.0cm]{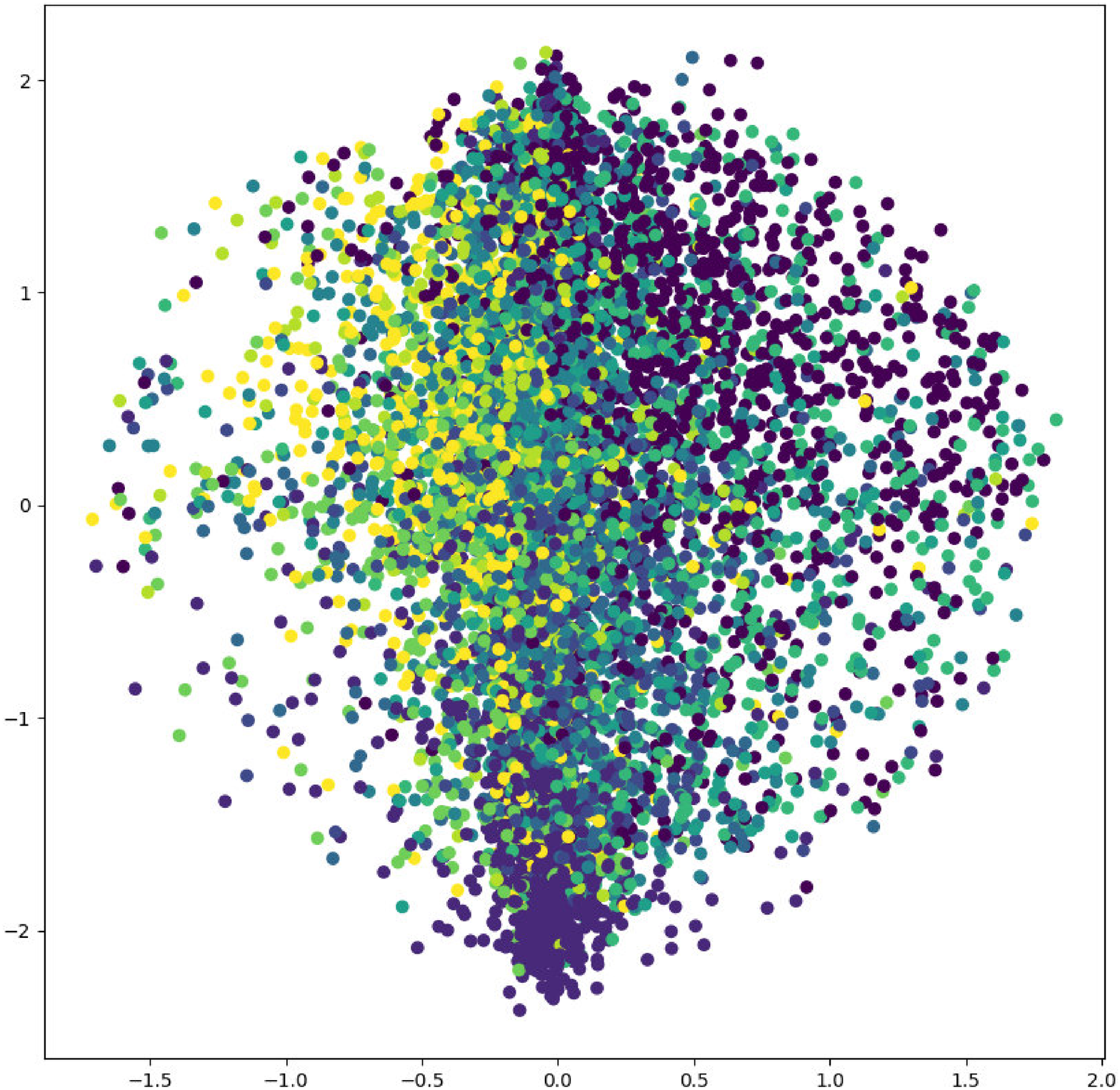}
\end{minipage}
}

\caption{The $\boldsymbol{\mu}$ and $\boldsymbol{z}$ latent spaces on USPS dataset. The figure clearly shows that compared with the $\boldsymbol{z}$ latent space, the $\boldsymbol{\mu}$ latent space has a better clusters pattern.}
\label{Pic-335-latent}
\end{figure*}

Although autoencoders can learn non-linear representations of the raw data, it turns out that the latent space of an autoencoder is compact and the clusters are prone to overlap with each other, leading to poor results when directly clustering on the latent space \cite{8622629}. Therefore, similar to their work, we deploy a deep dense neural network $DeepDenseNet_{\boldsymbol{\xi}}$ to transform the latent space of FAE to a new embedding space that encourages clear separations among different clusters. Specifically, we measure both the similarities in latent space and embedding space by probabilities. Because the VAE is a generative model, the latent variable $\boldsymbol{z}$ is confined by the spherical Gaussian distribution and gaps among different clusters in the latent space $\boldsymbol{z}$ are eliminated for a smooth transition during the generating process, which could cause severe overlapping between different clusters \cite{10.5555/3172077.3172161, 9207523, 8957256} and therefore hinder the deep dense neural network to separate different clusters. Thus we choose $\boldsymbol{\mu}$ instead of $\boldsymbol{z}$ to calculate the probabilities. To demonstrate that clusters in the $\boldsymbol{\mu}$ latent space have a more separate pattern compared with $\boldsymbol{z}$, clusters of these two latent spaces are manifested in Fig. \ref{Pic-335-latent} by using t-SNE \cite{maaten2008visualizing}. The distribution that measures the pairwise similarities in the latent space of FAE is defined as:

\begin{equation}\label{Eq-274-9}
P_{ij} = \frac{\left(1 + \left\Vert f(\boldsymbol{x}^{(i)}) - f(\boldsymbol{x}^{(j)}) \right\Vert^2 / \rho \right)^{-\frac{\rho + 1}{2}}}{\sum\nolimits_{i',j'} \left(1 + \left\Vert f(\boldsymbol{x}^{(i')}) - f(\boldsymbol{x}^{(j')}) \right\Vert^2 / \rho \right)^{-\frac{\rho + 1}{2}}}
\end{equation}

\noindent where $f(\boldsymbol{x}^{(i)}) = \boldsymbol{\mu}^{(i)}$ is the mean vector of data point $\boldsymbol{x}^{(i)}$ obtained from the inference model of FAE and $\rho$ is the number of degrees of freedom. Note that we construct the joint probabilities $P_{ij}$ through normalizing over all pairs of data samples. On the other hand, pairwise similarities in the embedding space are defined using the $t$-distribution with the degree of freedom $1$:

\begin{equation}\label{Eq-280-10}
Q_{ij} = \frac{\left(1 + \left\Vert g(\boldsymbol{\mu}^{(i)}) - g(\boldsymbol{\mu}^{(j)}) \right\Vert^2 \right)^{-1}}{\sum\nolimits_{i',j'} \left(1 + \left\Vert g(\boldsymbol{\mu}^{(i')}) - g(\boldsymbol{\mu}^{(j')}) \right\Vert^2 \right)^{-1}}
\end{equation}

\noindent in which $g(\boldsymbol{\mu}^{(i)}) = \boldsymbol{c}^{(i)}$ is the embedded vector of $\boldsymbol{\mu}^{(i)}$ yielded by the deep dense neural network. According to \cite{pmlr-v5-maaten09a}, a lower value for the degree of freedom could produce larger separation between clusters, hence we choose $1$ as the degree of freedom for the $Q$ distribution. Analogous to \cite{8622629}, the parameters of the deep dense neural network $\boldsymbol{\xi}$ are optimized by minimizing the cross-entropy between the distributions $P$ and $Q$:

\begin{equation}\label{Eq-286-11}
\mathcal{L}_{\gamma} = - \sum\nolimits_{i, j = 1}^M P_{ij} \mathop{log} \left( Q_{ij} \right).
\end{equation}

Finally, with all of these ingredients stated above, the objective function of the DCFAE is formulated as below:

\begin{subequations}
\begin{align}
\begin{split}
\mathop{min}_{\boldsymbol{\phi}, \boldsymbol{\theta}, \boldsymbol{\xi}} \mathcal{L}_{\alpha-\gamma} &= \mathcal{L}_\alpha + {\lambda}' \mathcal{L}_{\gamma} \\ &= \begin{aligned}[t] 
 &- \mathbb{E}_{q_{\boldsymbol{\phi}} (\boldsymbol{z}\ \vert\ \boldsymbol{x})} p_{\boldsymbol{\theta}} (\boldsymbol{x}\ \vert\ \boldsymbol{z}) + \mathcal{D}_{KL} \left( q_{\boldsymbol{\phi}} (\boldsymbol{z}\ \vert\ \boldsymbol{x})\ \Vert\ p(\boldsymbol{z}) \right) \\ &- \lambda\ \mathbb{E}_{q_{\boldsymbol{\phi}} (\boldsymbol{z}\ \vert\ \boldsymbol{x})} \mathop{log} D_{\boldsymbol{\psi}} (p_{\boldsymbol{\theta}} (\boldsymbol{x}\ \vert\ \boldsymbol{z})) - {\lambda}' \sum\nolimits_{i, j = 1}^M P_{ij} \mathop{log} \left( Q_{ij} \right),
 \end{aligned}
\end{split}
\label{Eq-294-12a} \\
\begin{split}
\mathop{min}_{\boldsymbol{\psi}} \mathcal{L}_\beta &= \mathcal{L}_D = - \frac{1}{2M} \sum\nolimits_{i = 1}^{2M} \tau_i \mathop{log} \tilde{\tau}_i + (1 - \tau_i) \mathop{log} (1 - \tilde{\tau}_i)
\end{split}
\label{Eq-295-12b}
\end{align}
\end{subequations}

\noindent where $\mathcal{L}_{\alpha-\gamma}$ indicates the combination of $\mathcal{L}_\alpha$ and $\mathcal{L}_\gamma$ which is balanced by parameter ${\lambda}'$.

\subsection{Neural network architecture}

All three components of FAE, i.e., the encoder, decoder, and the discriminator are built using deep residual convolutional neural networks. The residual block contains two convolutional layers with kernel size set as $3\times3$ and stride fixed to $1$. In the encoder network, we implement each convolutional layer using $2\times2$ stride for down-sampling, which is immediately followed by a ReLU activation layer. Then a residual block with the same number of filters is attached after the activation layer. The encoder contains $4$ groups of this combination and the outputs of the last group are flattened to connect with two dense layers that produce $\boldsymbol{\mu}$ and $\mathop{log} \boldsymbol{\sigma}$ separately. The network architecture of the discriminator is akin to the encoder network, except that the discriminator has a simpler layer combination and the output of the final dense layer is a single value that indicates the genuineness of an input image. The decoder basically possesses a reverse architecture of the encoder and we choose the transposed convolutional layer with $2\times2$ stride for up-sampling. Note that all transposed convolutional layers except the last one are followed by ReLU activation layers. 

\begin{figure*}[!ht]
\centering
\subfigure{
\begin{minipage}[t]{3\textwidth}
\includegraphics[height=9.27cm,width=16.5cm]{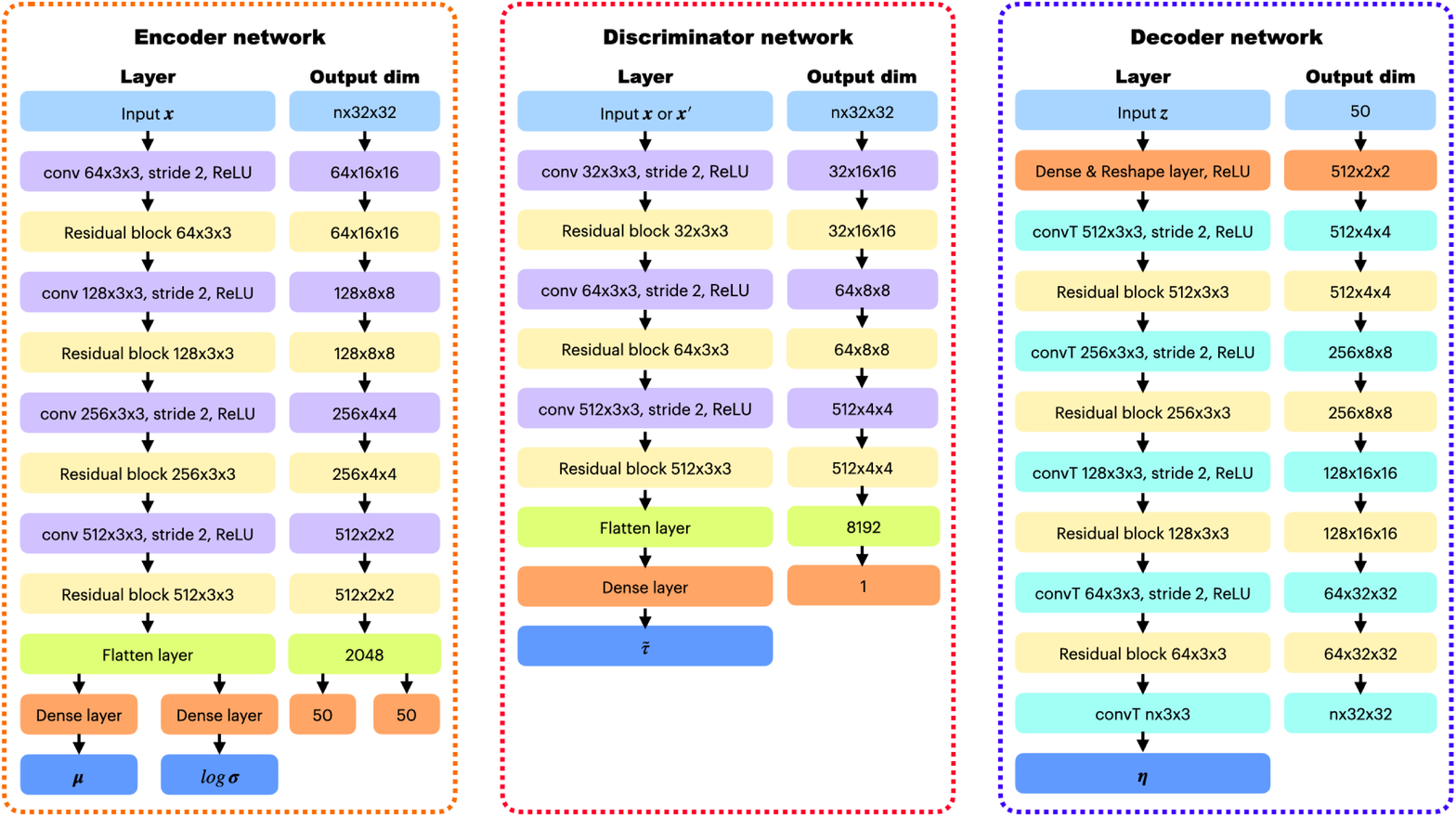}
\end{minipage}
}

\subfigure{
\begin{minipage}[t]{0.85\textwidth}
\includegraphics[height=8.92cm,width=14cm]{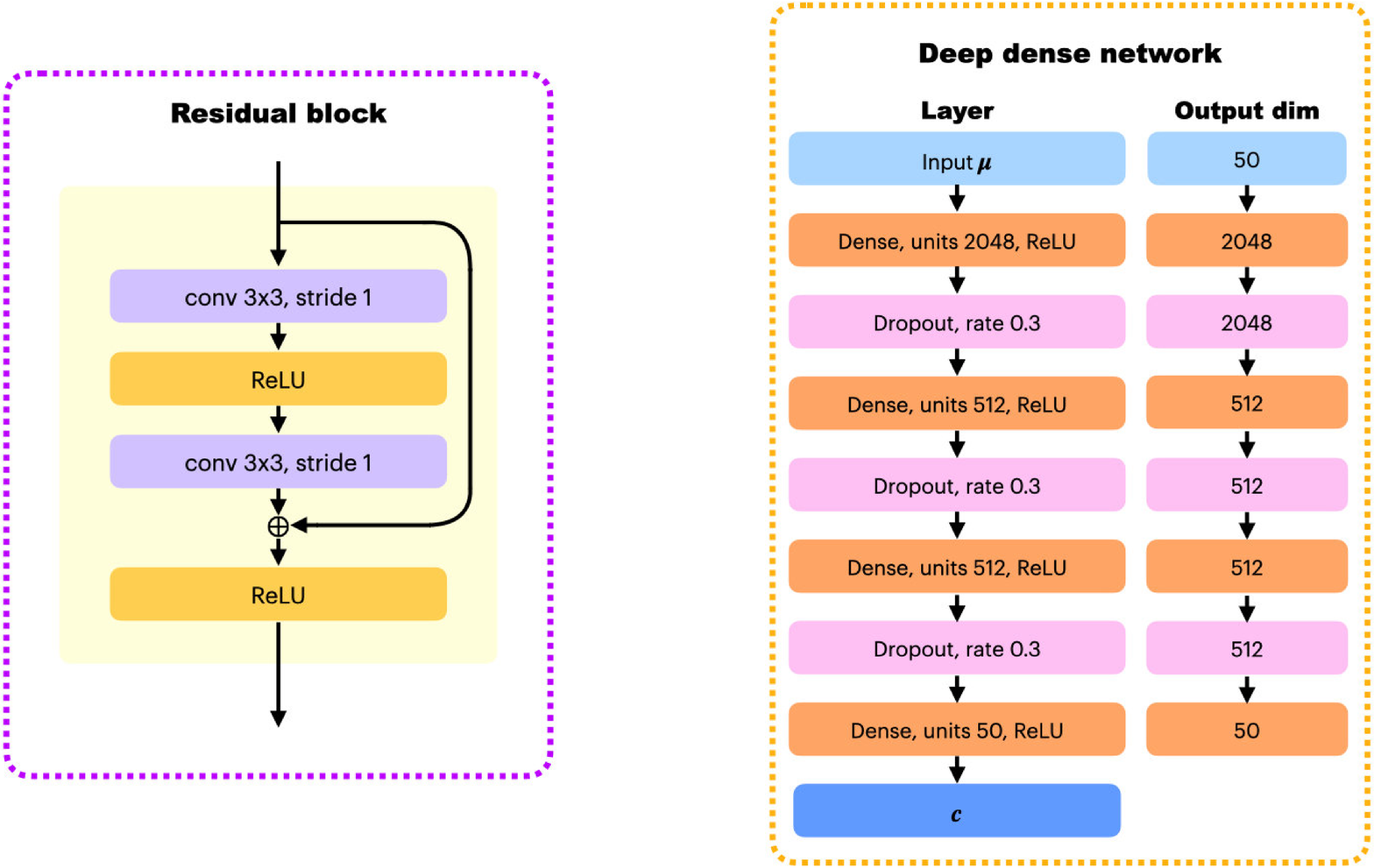}
\end{minipage}
}
\caption{Neural network architecture of the DCFAE model}
\label{Pic-316-1}
\end{figure*}

The deep dense neural network is constructed by stacking $4$ dense layers with the ReLU activation function. In addition, we add $L2$ weight regularization to all dense layers and attach a dropout layer after each of the first $3$ dense layers for deterring overfitting. Skeletons of all the network architectures are depicted in Fig. \ref{Pic-316-1}.

\subsection{Optimization process}

The deep neural network optimization strategy demonstrated in \cite{hinton2006reducing} is witness to the efficiency of using a pre-trained initialization for fine-tuning the deep networks. Analogically, we divide the optimization process of DCFAE into two phases, viz. the pre-train phase and the fine-tuning phase. In the pre-train phase, the deep dense neural network $DeepDenseNet_{\boldsymbol{\xi}}$ is frozen and only the FAE model is trained. After the pre-training, all parts of the DCFAE model are trained in the fine-tuning phase. The whole optimization procedure is sketched in Algorithm \ref{Algo-357-1}.

\begin{algorithm}
\caption{Algorithm for optimizing Deep Clustering with Fusion AutoEncoder (DCFAE)}
\label{Algo-357-1}
\KwIn {The original data $\mathcal{X} = \left\{\boldsymbol{x}^{(1)}, \boldsymbol{x}^{(2)}, \ldots, \boldsymbol{x}^{(n)}\right\}$, parameters $\lambda, \ \gamma$, cluster number $clusters$, batch size $M$, execute iterations $epochs$.}
\KwOut {Network parameters of DCFAE $\boldsymbol{\phi}$, $\boldsymbol{\theta}$, $\boldsymbol{\psi}$, and $\boldsymbol{\xi}$, clustering label of $\boldsymbol{x}^{i}$ $c_i$.}
\textbf{Initialize:} \\
Initialize network parameters $\boldsymbol{\phi}$, $\boldsymbol{\theta}$, and $\boldsymbol{\psi}$ of the FAE model through the pre-train phase. \\
\textbf{Optimizing:}\\
   \For{$epoch$ in $epoches$}{
      \For{every $i \in \left\{ 1, 2, \ldots, \left\lfloor \frac{n}{M} \right\rfloor \right\}$} {
          Sample a batch of training data $\mathcal{X}^{(i)}$ from $\mathcal{X}$\;
          Forward pass $\mathcal{X}^{(i)}$ to the encoder to yield $\boldsymbol{\mu}$ and $\mathop{log} \boldsymbol{\sigma}$, then compute the latent representations $\boldsymbol{z}$ using Eq. \eqref{Eq-232-5}\;
          Forward pass $\boldsymbol{z}$ to the decoder to produce the reconstruction $\widetilde{\mathcal{X}^{(i)}}$, then compute the $\mathcal{L}_{ELBO}$ loss using Eq. \eqref{Eq-220-4}\;
          Construct the double-batch size label $\boldsymbol{\tau}$ and data $\left[\widetilde{\mathcal{X}^{(i)}}, \mathcal{X}^{(i)}\right]$, then forward pass $\left[\widetilde{\mathcal{X}^{(i)}}, \mathcal{X}^{(i)}\right]$ to the discriminator to get the predicted label $\tilde{\boldsymbol{\tau}}$\;
          Forward pass $\widetilde{\mathcal{X}^{(i)}}$ to discriminator to get the predicted label $\boldsymbol{\tau}'$\;
          Compute the losses $\mathcal{L}_D$ and $\mathcal{L}_G$ according to Eqs. \eqref{Eq-243-6} and \eqref{Eq-250-7}, respectively\;
          Forward pass $\boldsymbol{\mu}$ to the deep dense net to get the embedding $\boldsymbol{c}$\;
          Compute the loss $\mathcal{L}_{\gamma}$ using Eq. \eqref{Eq-286-11}\;
          Update the DCFAE model parameters with Eqs. \eqref{Eq-294-12a} and \eqref{Eq-295-12b} in sequence.
      }
       }
Forward pass $\mathcal{X}$ to the encoder to get $\boldsymbol{\mu}$\;
Forward pass $\boldsymbol{\mu}$ to the deep dense net to yield $\boldsymbol{c}$\;
Perform $k$-means algorithm on the rows of $\boldsymbol{c}$ to get the final clustering result.
\end{algorithm}

\section{Experiments}

This section presents the experiments that assess the effectiveness and stability of the DCFAE model, including the comparison experiment, ablation study, adversarial training convergence experiment, and parameter analysis.

\subsection{Implementation details}

DCFAE is implemented using the Tensorflow (Version 2.4.1) machine learning library. All experiments are conducted on a Dell workstation with Windows 10 Pro (Version 21H1) installed. It is equipped with 32 GB 2666 MHz ECC memory, an intel$^{\circledR}$ Xeon$^{\circledR}$ W-2135 $@$ 3.70 GHz CPU, and an Nvidia GeForce$^{\circledR}$ GTX 1080 Ti GPU with 11 GB GDDR5X video memory.

Key experimental configurations are listed as follows: the batch size is 256, the dropout rate of the dropout layer is 0.3, Adam algorithm \cite{DBLP:journals/corr/KingmaB14} is adopted as the optimizer and the learning rate is ${10}^{-4}$. The dimension of the latent variable $L$, the number of degrees of freedom $\rho$, and the balance parameters $\lambda$ and ${\lambda}'$ are set as 50, 100, 100, and 10, respectively. In addition, we resize all images to $32\times32$ and exploit the data augmentation strategies including rotation as well as width and height shift for increasing the diversity of the training set, which is useful to stabilize the adversarial training process.

\subsection{Datasets}

We choose 5 image datasets to assess the clustering performance of our proposed model. Details about these datasets are given below.

$\bm{MNIST}$ and $\bm{MNIST}$-$\bm{test}$ are large handwritten digits datasets which consist of 0-9 human written digits manifested in $28\times28$ gray-scale images. There are 60,000 training samples and 10,000 test samples in the MNIST dataset, and we combine all the images as a whole dataset for the following experiments. The MNIST-test dataset is the test fraction of MNIST and therefore contains 10,000 datapoints. Both of these two datasets have 10 separate clusters.

$\bm{USPS}$ is composed of 9,298 $16\times16$ gray-scale digit images scanned from envelopes. Every image depicts one specific digit of 0-9 and therefore the dataset has 10 different classes.

$\bm{Fashion}$-$\bm{MNIST}$ shares similar specifications with the MNIST dataset except that the digits are replaced by 10 categories of fashion products, such as T-shirts, coats, bags, etc. We unite the training and test sets for evaluation.

$\bm{COIL}$-$\bm{100}$ contains 7200 $128 \times 128$ color images of 100 objects. Pictures of every object are captured by rotating 5 degrees each time against a fixed color camera, yielding 72 poses per object.

The specifications of these datasets are summarized in Table \ref{Tb-391-1}.

\begin{table*}[!h]
\centering
\fontsize{9}{15}\selectfont
\setlength{\tabcolsep}{7pt}
\setlength{\arrayrulewidth}{0.3mm}
\caption{\fontsize{10pt}{\baselineskip}\selectfont A summary of the testing datasets.}
\begin{tabular}{c c c c} 
 \hline
 Dataset & \# Instance & \# Cluster & \# Dimension \\ [0.5ex] 
\hline
 MNIST & 70,000 & 10 & $28\times28\times1$ \\ 

 MNIST-test & 10,000 & 10 & $28\times28\times1$ \\

 USPS & 9298 & 10 & $16\times16\times1$ \\
  
 Fashion-MNIST & 70,000 & 10 & $28\times28\times1$ \\ 
 
 COIL-100 & 7200 & 100 & $128\times128\times3$ \\
 \hline
\end{tabular}
\label{Tb-391-1}
\end{table*}

\subsection{Baseline models and evaluation measures}

Several relative clustering models are picked for comparing the clustering performance against DCFAE. We first choose k-means as a classical model and then select 7 DC methods for the comparison. These DC models can be split into three categories: the autoencoder based camp includes \textbf{DEC} \cite{xie2016unsupervised}, \textbf{DEPICT} \cite{Dizaji_2017_ICCV}, and \textbf{LECDR} \cite{8622629}; generative model based methods \textbf{VaDE} \cite{10.5555/3172077.3172161}, \textbf{DC-VAE} \cite{XU2020106260}, and \textbf{ClusterGAN} \cite{8954410}; while \textbf{JULE} \cite{7780925} belongs to the direct cluster optimization DC models. 

Two popular clustering performance measures are adopted for the comparison judgment, i.e., the accuracy (\textbf{ACC}) and the normalized mutual information (\textbf{NMI}). Details of both metrics can be found in \cite{xu2003document} and higher values indicate better clustering performance. We execute DCFAE 10 times and record the mean values for the comparison. 

\subsection{Comparison results}

The comparison results of all clustering algorithms are list in Table \ref{Tb-415-2}. The symbol `$-$' means the clustering results can be obtained neither from the original paper nor there are compatible codes for running.

From the table, we can find that DCFAE performs well on all datasets and surpasses all other methods in most cases, which confirms the effectiveness of our proposed model. In addition, DCFAE gets the highest scores among the Average columns, suggesting it has better compatibility when comes to different datasets. Note that compared with the VaDE and DC-VAE model, DCFAE possesses superior clustering performance in most scenarios, which indirectly proves the improvements of FAE to the VAE model. Since our designed algorithm shares a similar clustering loss objective with LECDR, the clustering results w.r.t. these two models demonstrate the validity of the proposed clustering strategy.

\begin{table*}[!h]
\centering
\fontsize{7.2}{13}\selectfont
\setlength{\tabcolsep}{9pt}
\setlength{\arrayrulewidth}{0.3mm}
\caption{\fontsize{10pt}{\baselineskip}\selectfont Clustering performance of each algorithm in terms of ACC and NMI on five image datasets.}
\begin{tabular}{l cc cc cc cc cc cc}
\toprule
 & \multicolumn{2}{c}{MNIST} & \multicolumn{2}{c}{MNIST-test} & \multicolumn{2}{c}{USPS} & \multicolumn{2}{c}{Fashion-MNIST} & \multicolumn{2}{c}{COIL-100} & \multicolumn{2}{c}{Average} \\
\cmidrule(lr){2-3} \cmidrule(lr){4-5} \cmidrule(lr){6-7} \cmidrule(lr){8-9} \cmidrule(lr){10-11} \cmidrule(lr){12-13}
Model & ACC & NMI & ACC & NMI & ACC & NMI & ACC & NMI & ACC & NMI & ACC &NMI \\
\midrule
k-means & 54.20 & 48.11 & 54.84 & 49.99 & 67.34 & 62.72 & 47.52 & 51.10 & 62.25 & 83.78 & 57.23 & 59.14 \\
JULE & 96.40 & 91.30 & 96.10 & 91.50 & 95.00 & 91.30 & 56.30 & 60.80 & 77.40 & 83.00 & 84.24 & 83.58\\
DEC & 87.93 & 80.71 & 85.60 & 83.00 & 70.86 & 68.40 & 60.61 & 63.70 & 75.50 & 79.00 & 76.10 & 74.96\\
DEPICT & 96.50 & 91.70 & 96.30 & 91.50 & 96.40 & 92.70 & 58.30 & 62.00 & 66.50 & 74.00 & 82.80 & 82.38\\
LECDR & 97.02 & 92.34 & $-$ & $-$ & 77.31 & 80.53 & 58.41 & 63.25 & $-$ & $-$ & 77.58 & 78.71\\
ClusterGAN & 96.40 & 92.10& $-$ & $-$ & 97.00 & \textbf{93.10} & 60.50 & 64.00 & \textbf{84.10} & 80.00 & 84.50 & 82.3\\
VaDE & 94.46 & 90.74 & 28.70 & 28.70 & 56.60 & 51.20 & 57.80 & 63.00 & $-$ & $-$ & 59.39 & 58.41\\
DC-VAE & 97.30 & \textbf{94.10} & $-$ & $-$ & 77.80 & 69.70 & 59.70 & 63.30 & $-$ & $-$ & 78.27 & 75.70\\
DCFAE & \textbf{97.36} & 93.29 & \textbf{97.56} & \textbf{93.70} & \textbf{97.30} & 93.03 & \textbf{64.34} & \textbf{68.26} & 82.79 & \textbf{91.39} & \textbf{87.87} & \textbf{87.93}\\
\bottomrule
\end{tabular}
\label{Tb-415-2}
\end{table*}

\subsection{Ablation study}

\subsubsection{Effectiveness of the FAE framework}

Since the FAE architecture is the main novelty of our work, an ablation experiment that strips off the discriminator is performed to inspect the representation learning ability of the FAE model. Specifically, we remove the discriminator of the FAE to form a VAE which is built with the deep residual convolutional network, and the deep dense neural network is attached to the VAE to learn the embedding space. The altered model is denoted as $DCFAE - Dis$ (`$-$' here means minus, the same notation is applied for the following ablation experiments) and the comparison results are listed in Table \ref{Tb-425-3}. Note that we add another clustering performance measurement adjusted rand index (\textbf{ARI}) \cite{JMLR:v11:vinh10a} in the following experiments.

\begin{table*}[!h]
\centering
\fontsize{7.2}{13}\selectfont
\setlength{\tabcolsep}{5.7pt}
\setlength{\arrayrulewidth}{0.3mm}
\caption{\fontsize{10pt}{\baselineskip}\selectfont Clustering results w.r.t. DCFAE - Dis and DCFAE}
\begin{tabular}{l ccc ccc ccc ccc ccc}
\toprule
 & \multicolumn{3}{c}{MNIST} & \multicolumn{3}{c}{MNIST-test} & \multicolumn{3}{c}{USPS} & \multicolumn{3}{c}{Fashion-MNIST} & \multicolumn{3}{c}{COIL-100}\\
\cmidrule(lr){2-4} \cmidrule(lr){5-7} \cmidrule(lr){8-10} \cmidrule(lr){11-13} \cmidrule(lr){14-16}
Model & ACC & NMI & ARI & ACC & NMI & ARI & ACC & NMI & ARI & ACC & NMI & ARI & ACC & NMI & ARI \\
\midrule
DCFAE - Dis & 95.81 & 90.67 & 90.99 & 95.97 & 91.03 & 91.37 & 96.16 & 91.23 & 92.76 & 62.25 & 65.92 & 50.51 & 80.39 & 90.87 & 75.39 \\
DCFAE & \textbf{97.36} & \textbf{93.29} & \textbf{94.26} & \textbf{97.56} & \textbf{93.70} & \textbf{94.70} & \textbf{97.30} & \textbf{93.03} & \textbf{94.72} & \textbf{64.34} & \textbf{68.26} & \textbf{52.40} & \textbf{82.79} & \textbf{91.39} & \textbf{77.10} \\
\bottomrule
\end{tabular}
\label{Tb-425-3}
\end{table*}

\begin{figure*}[!ht]
\centering
\subfigure[Original images]{
\begin{minipage}[t]{0.31\textwidth}
\includegraphics[height=5.0cm,width=5.0cm]{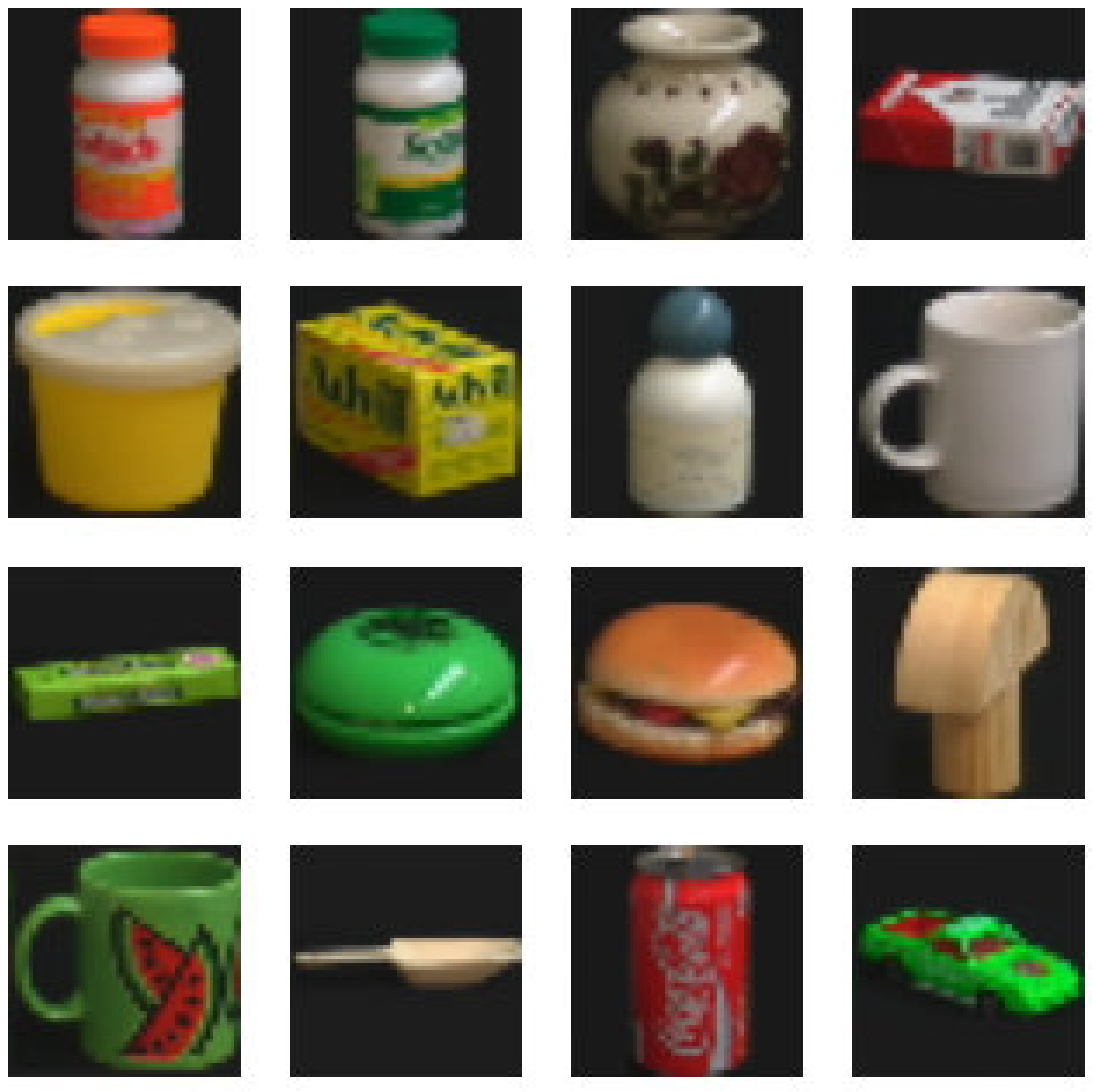}
\end{minipage}
}
\subfigure[Images reconstructed by DCFAE - Dis]{
\begin{minipage}[t]{0.31\linewidth}
\includegraphics[height=5.0cm,width=5.0cm]{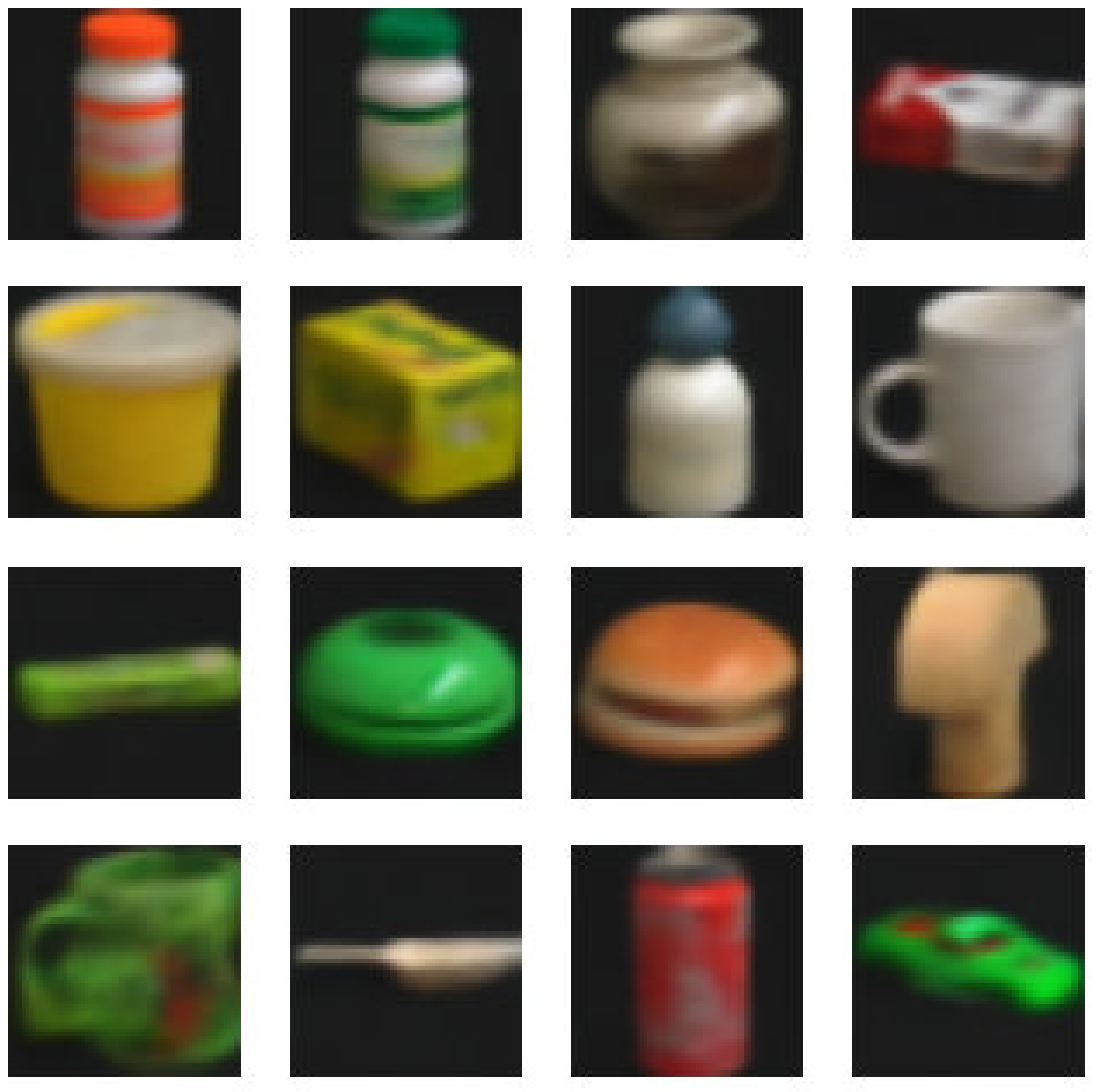}
\end{minipage}
}
\subfigure[Sampled images from DCFAE - Dis]{
\begin{minipage}[t]{0.31\linewidth}
\includegraphics[height=5.0cm,width=5.0cm]{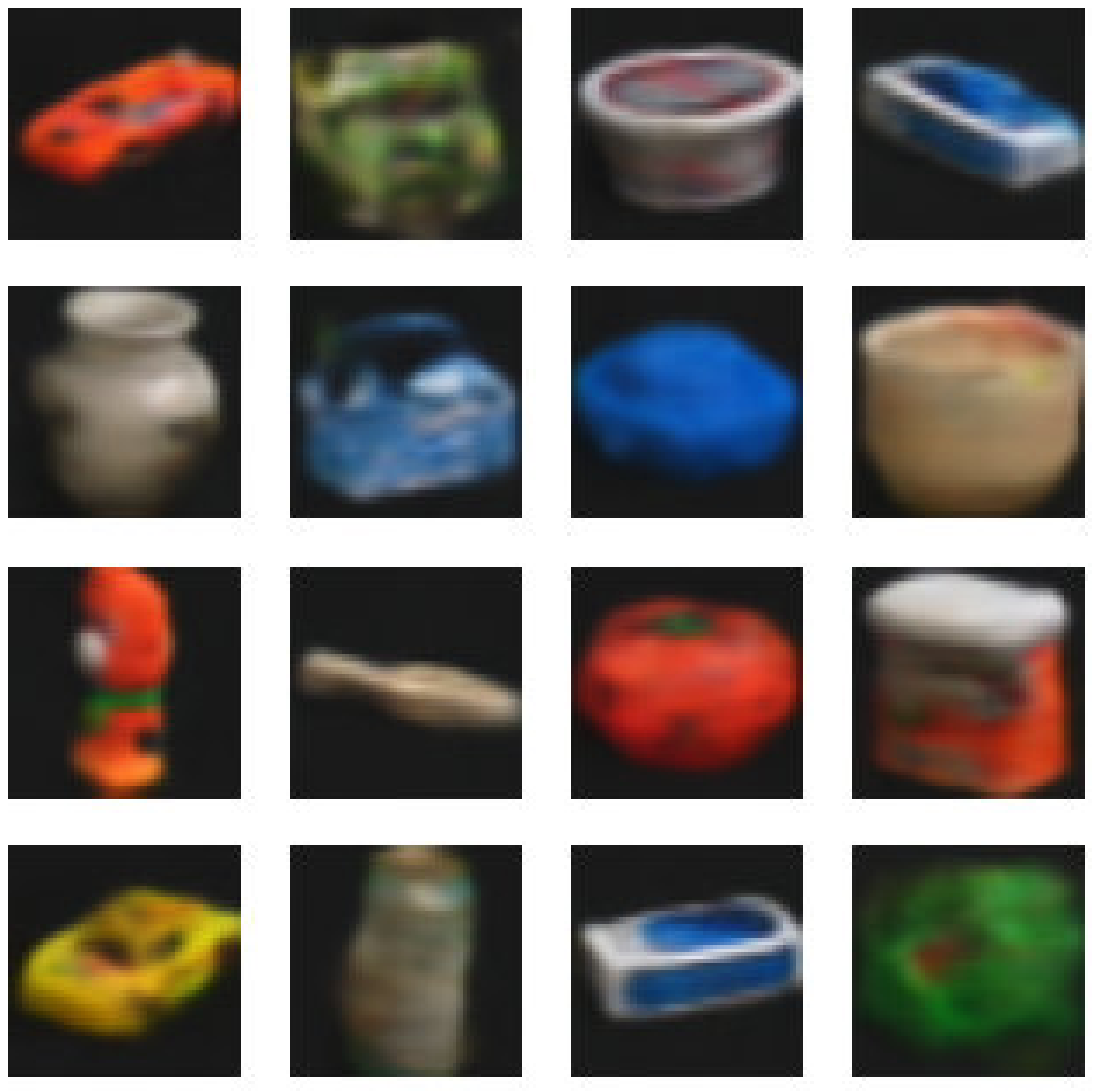}
\end{minipage}
}

\subfigure[Original images]{
\begin{minipage}[t]{0.31\linewidth}
\includegraphics[height=5.0cm,width=5.0cm]{coil_original.eps}
\end{minipage}
}
\subfigure[Images reconstructed by DCFAE]{
\begin{minipage}[t]{0.31\linewidth}
\includegraphics[height=5.0cm,width=5.0cm]{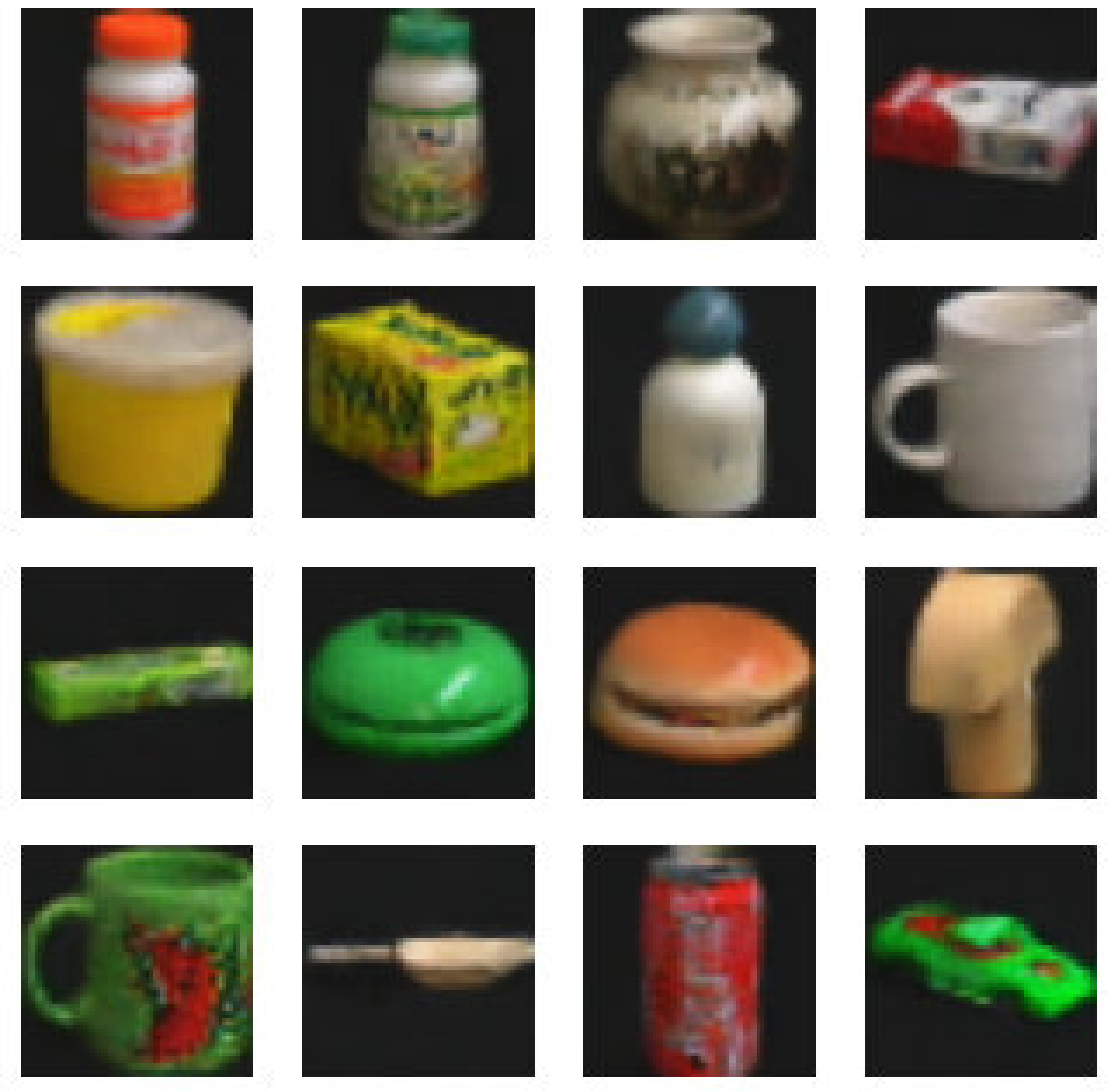}
\end{minipage}
}
\subfigure[Sampled images from DCFAE]{
\begin{minipage}[t]{0.31\linewidth}
\includegraphics[height=5.0cm,width=5.0cm]{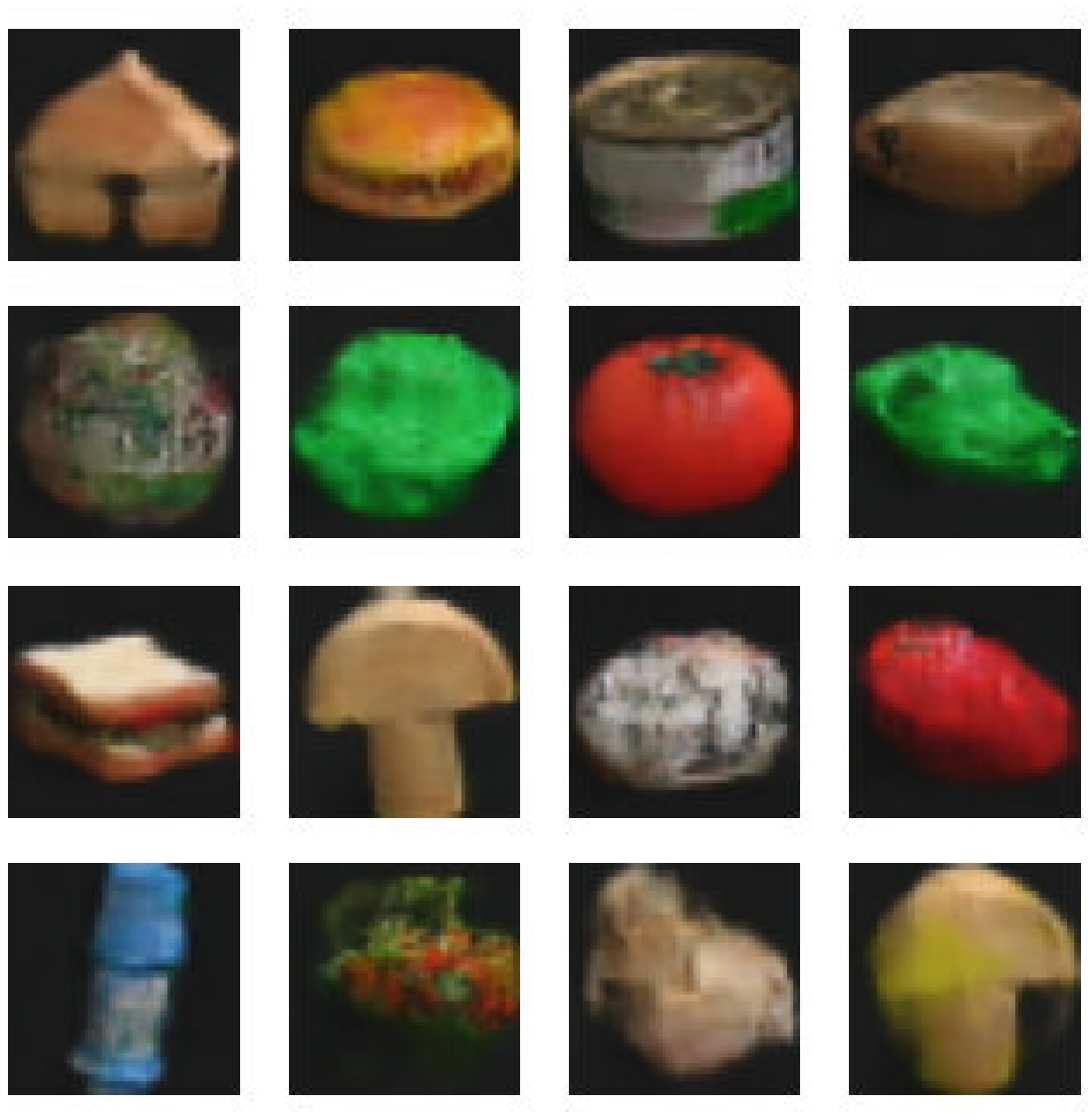}
\end{minipage}
}

\caption{Original, reconstructed, and sampled images. (a) and (d) are the same original images that fed into the DCFAE - Dis or DCFAE model; (b) and (e) are images that reconstructed by the DCFAE - Dis and DCFAE model, respectively; (c) and (f) are images that randomly sampled from the DCFAE - Dis and DCFAE model, respectively. It is obvious that compared with the DCFAE - Dis model, DCFAE definitely produces clear images regardless of reconstruction or sampling.}
\label{Pic-500-ab-1}
\end{figure*}

Table \ref{Tb-425-3} points out that DCFAE has better clustering performance on all 5 datasets in contrast to the DCFAE - Dis model, implying that the amalgamation of VAE and GAN can actually capture the subtle discriminative latent features which are helpful for the downstream clustering task. In addition, Fig. \ref{Pic-500-ab-1} demonstrates that the FAE framework could effectively alleviate the image blurriness issue of the VAE model.

\subsubsection{Effectiveness of the deep residual convolutional neural network architecture}

To investigate the contribution of the deep residual convolutional neural network architecture to the proposed model, we drop all the residual blocks in the network architecture of FAE and keep other configurations unchanged. The modified clustering model is referred to as DCFAE - Resnet and the clustering results are shown in Table \ref{Tb-477-4}.

\begin{table*}[!h]
\centering
\fontsize{7.2}{13}\selectfont
\setlength{\tabcolsep}{5.7pt}
\setlength{\arrayrulewidth}{0.3mm}
\caption{\fontsize{10pt}{\baselineskip}\selectfont Clustering results w.r.t. DCFAE - Resnet and DCFAE}
\begin{tabular}{l ccc ccc ccc ccc ccc}
\toprule
 & \multicolumn{3}{c}{MNIST} & \multicolumn{3}{c}{MNIST-test} & \multicolumn{3}{c}{USPS} & \multicolumn{3}{c}{Fashion-MNIST} & \multicolumn{3}{c}{COIL-100}\\
\cmidrule(lr){2-4} \cmidrule(lr){5-7} \cmidrule(lr){8-10} \cmidrule(lr){11-13} \cmidrule(lr){14-16}
Model & ACC & NMI & ARI & ACC & NMI & ARI & ACC & NMI & ARI & ACC & NMI & ARI & ACC & NMI & ARI \\
\midrule
DCFAE - Resnet & 95.52 & 89.54 & 90.40 & 96.31 & 91.32 & 92.02 & 96.03 & 90.44 & 92.23 & 56.34 & 62.72 & 46.83 & \textbf{86.18} & \textbf{93.70} & \textbf{81.98} \\
DCFAE & \textbf{97.36} & \textbf{93.29} & \textbf{94.26} & \textbf{97.56} & \textbf{93.70} & \textbf{94.70} & \textbf{97.30} & \textbf{93.03} & \textbf{94.72} & \textbf{64.34} & \textbf{68.26} & \textbf{52.40} & 82.79 & 91.39 & 77.10 \\
\bottomrule
\end{tabular}
\label{Tb-477-4}
\end{table*}

\begin{figure*}[!ht]
\centering
\subfigure[Original images]{
\begin{minipage}[t]{0.31\textwidth}
\includegraphics[height=5.0cm,width=5.0cm]{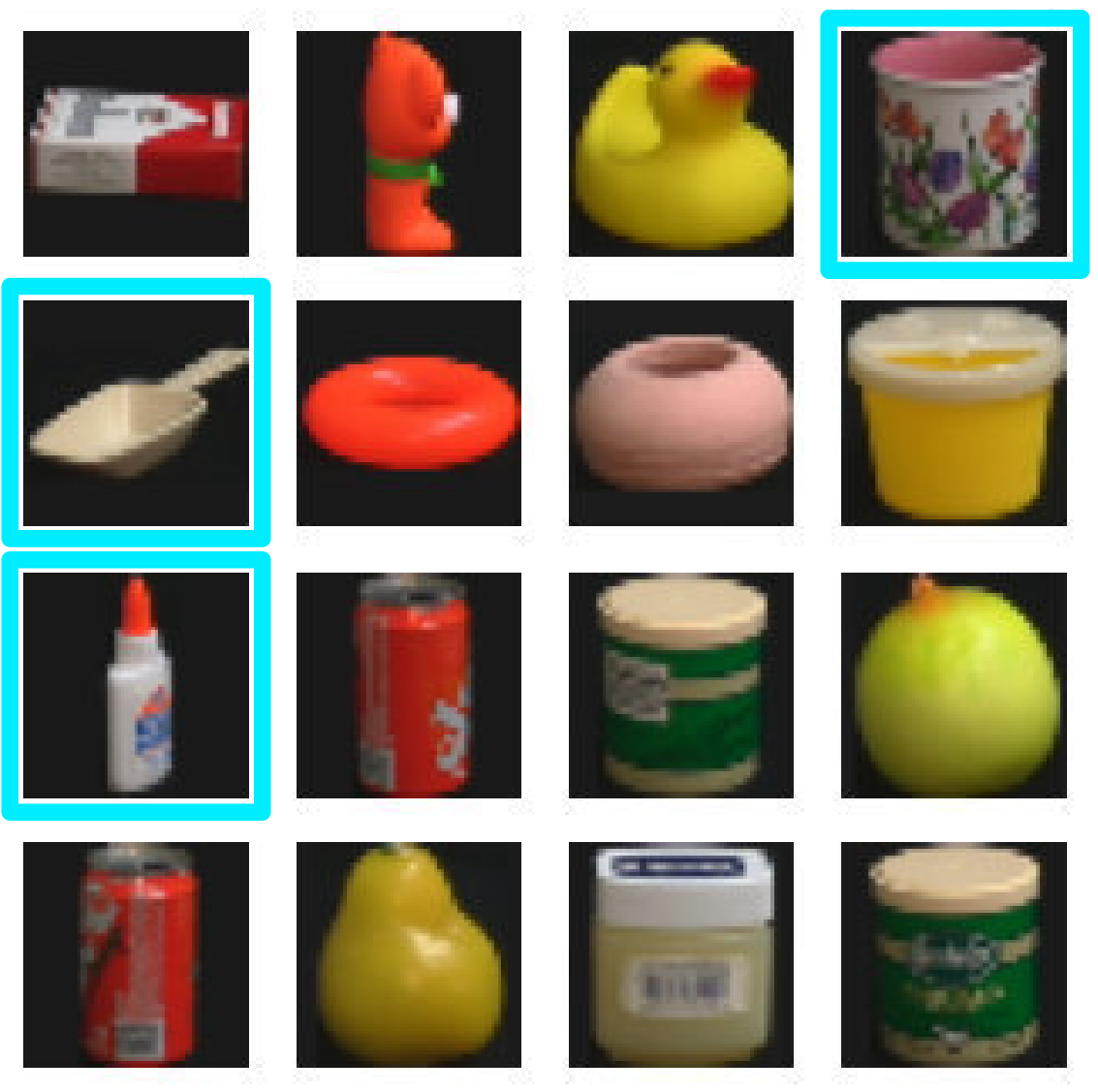}
\end{minipage}
}
\subfigure[Images reconstructed by DCFAE - Resnet]{
\begin{minipage}[t]{0.31\linewidth}
\includegraphics[height=5.0cm,width=5.0cm]{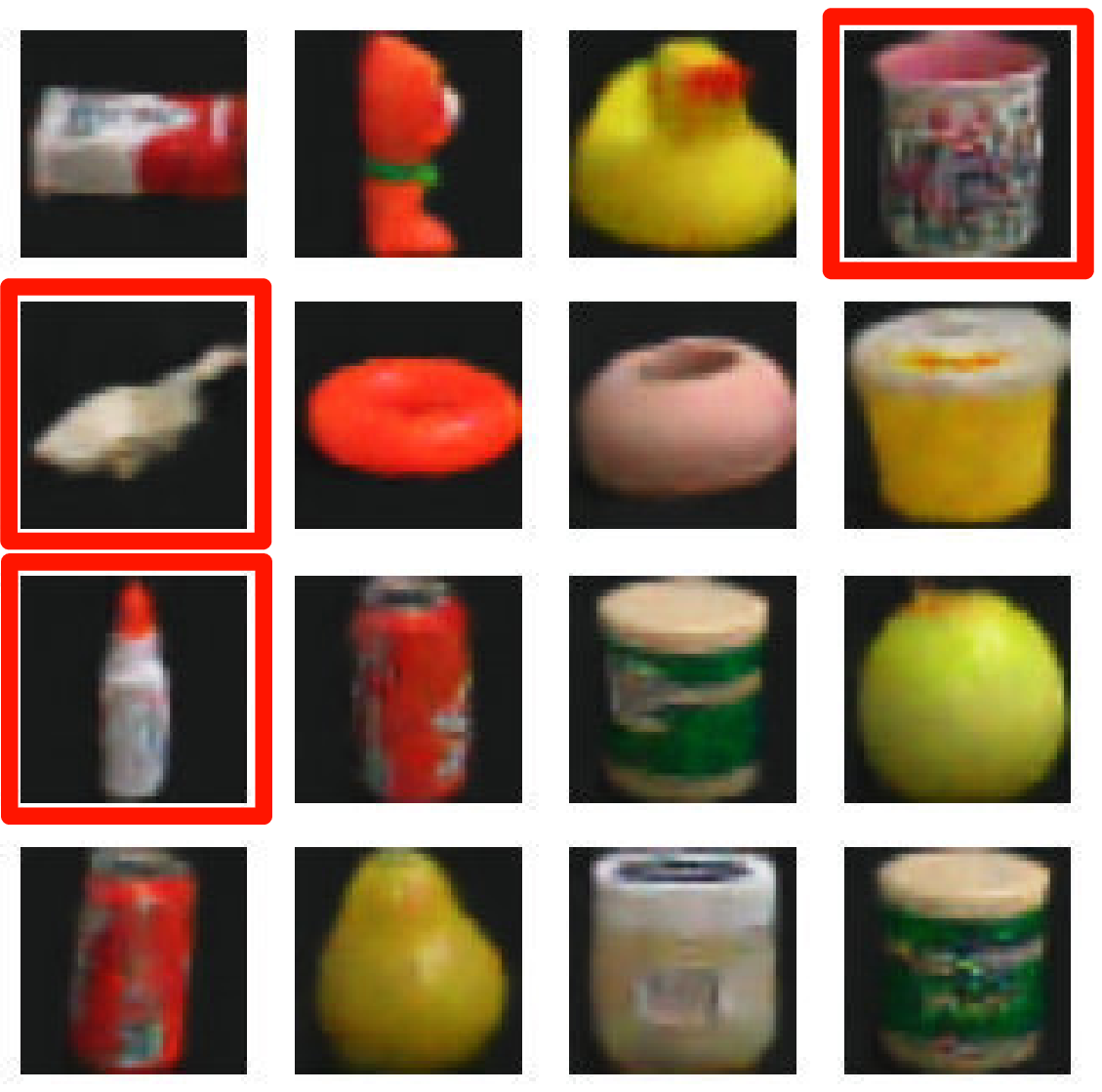}
\end{minipage}
}
\subfigure[Sampled images from DCFAE - Resnet]{
\begin{minipage}[t]{0.31\linewidth}
\includegraphics[height=5.0cm,width=5.0cm]{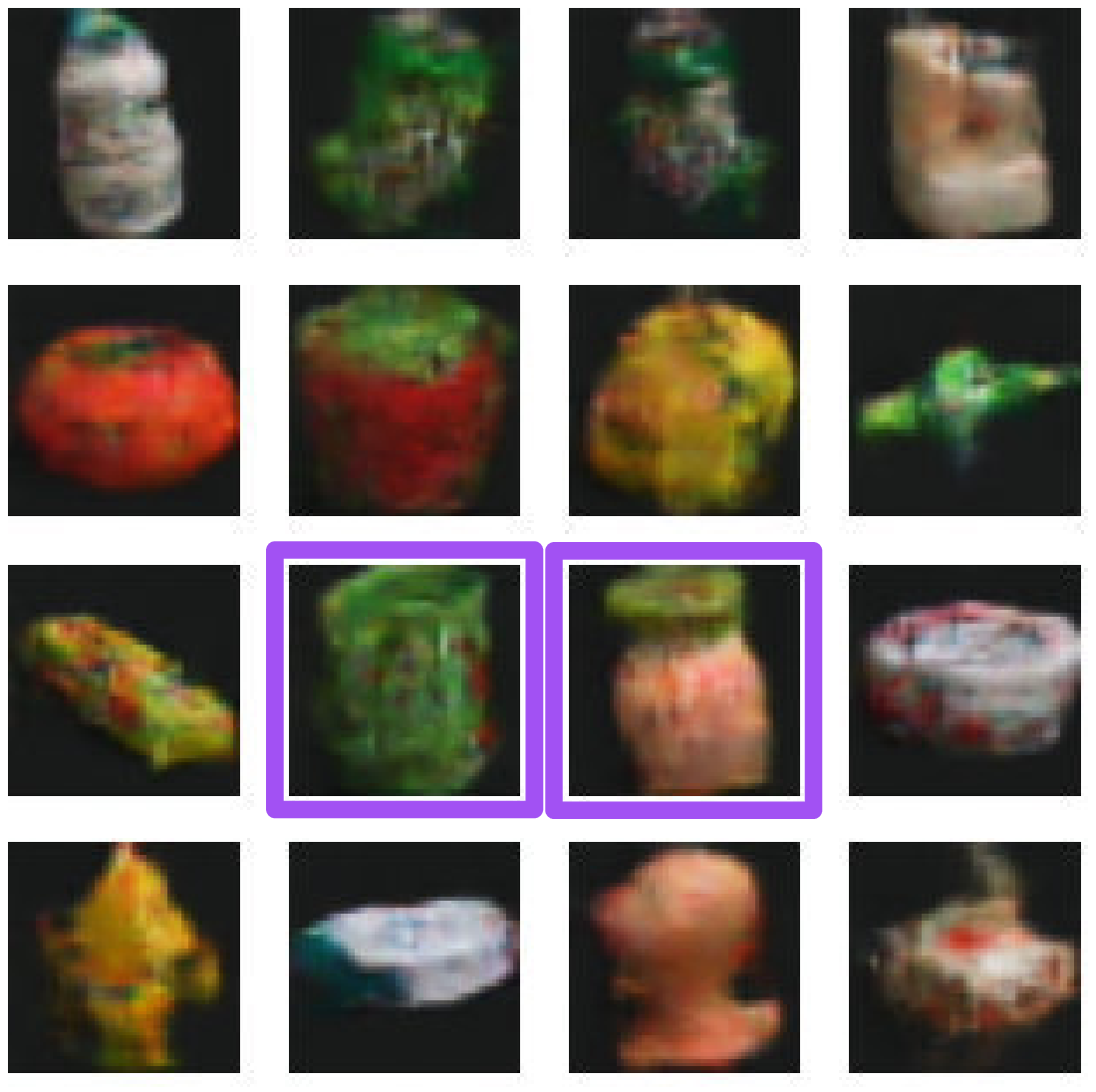}
\end{minipage}
}

\subfigure[Original images]{
\begin{minipage}[t]{0.31\linewidth}
\includegraphics[height=5.0cm,width=5.0cm]{coil_ab_2_original.eps}
\end{minipage}
}
\subfigure[Images reconstructed by DCFAE]{
\begin{minipage}[t]{0.31\linewidth}
\includegraphics[height=5.0cm,width=5.0cm]{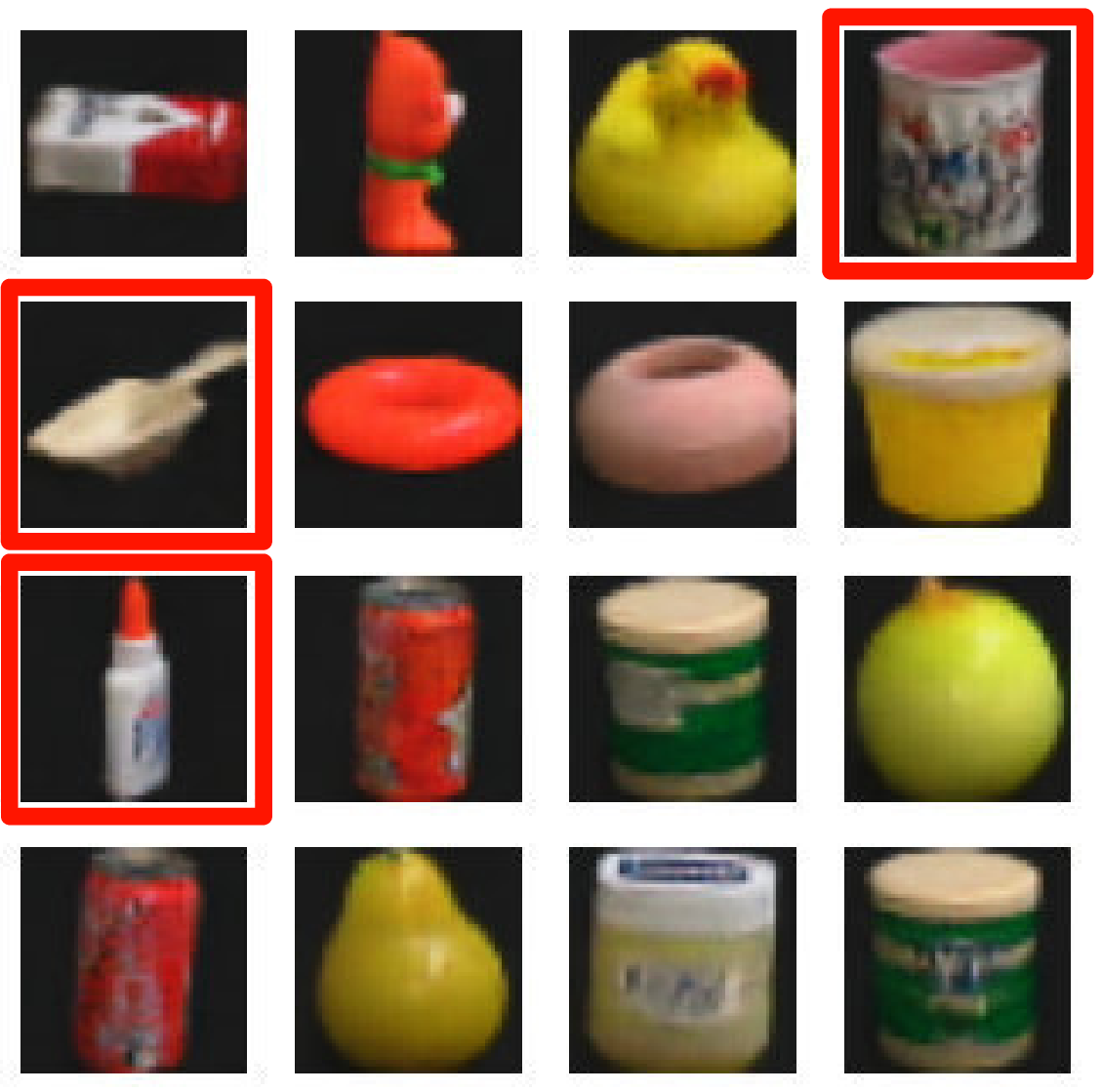}
\end{minipage}
}
\subfigure[Sampled images from DCFAE]{
\begin{minipage}[t]{0.31\linewidth}
\includegraphics[height=5.0cm,width=5.0cm]{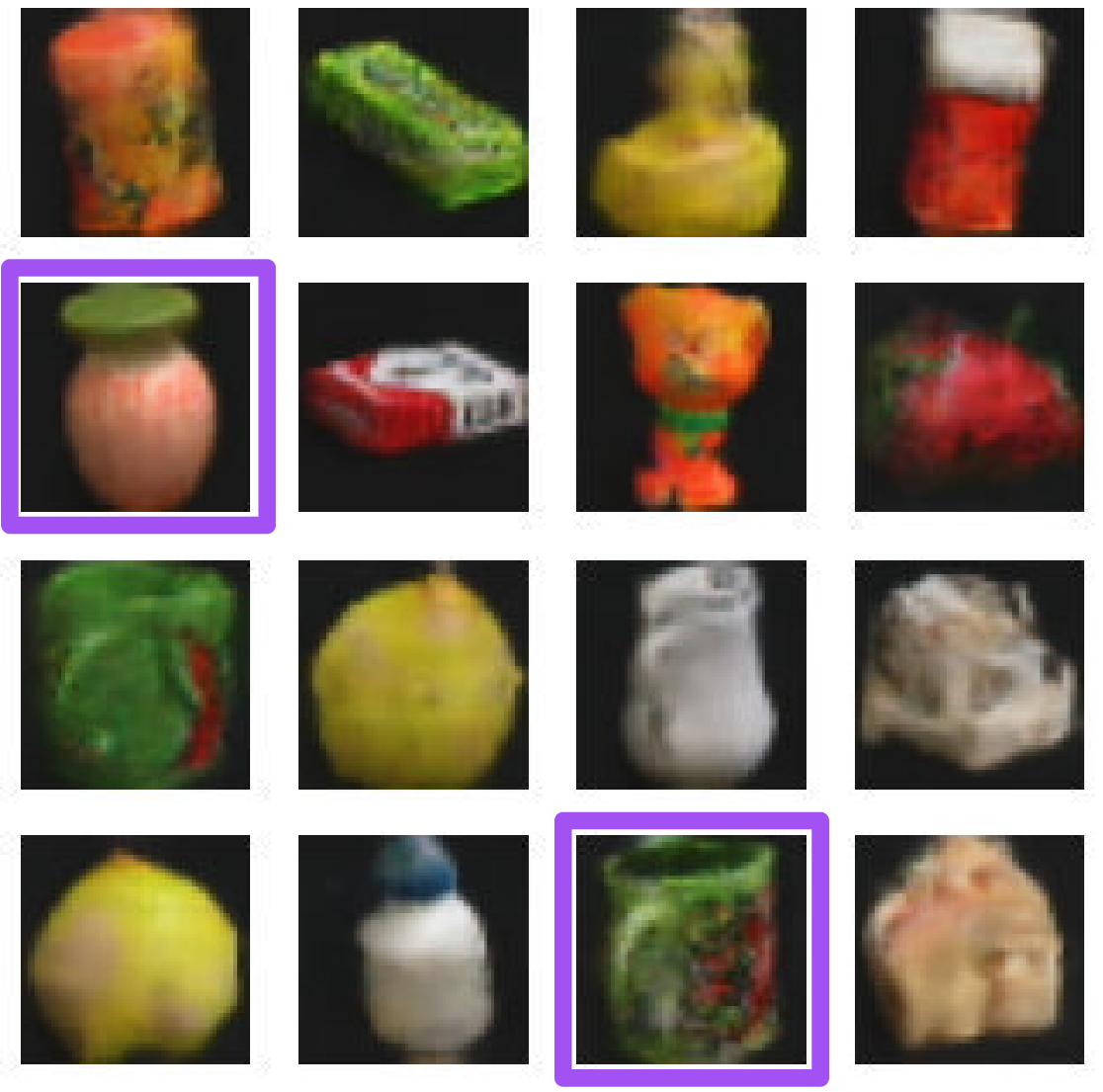}
\end{minipage}
}

\caption{Original, reconstructed, and sampled images. (a) and (d) are the same original images that fed into the DCFAE - Resnet or DCFAE model; (b) and (e) are images that reconstructed by the DCFAE - Resnet and DCFAE model, respectively; (c) and (f) are images that randomly sampled from the DCFAE - Resnet and DCFAE model, respectively. Note that in the reconstructed images, we rim some images (the corresponding images are also rimmed in the original images by the cyan frames) using red frames to highlight that DCFAE can more accurately reconstruct the input images compared with DCFAE - Resnet. Besides, it is clear that images sampled by DCFAE have better qualities and we rim two objects by purple frames in (c) and (f) to emphasize the discrepancies.}
\label{Pic-523}
\end{figure*}

As the table suggested, adding residual blocks to the FAE framework could increase the cluster results on most of the datasets, which indicates that FAE can benefit from the deep residual convolutional network architecture in terms of representation learning. Note that there is a clustering performance bump when performing the ablation on the COIL-100 dataset. Nevertheless, Fig. \ref{Pic-523} shows that the FAE will produce higher quality reconstructed and generated images when built with the deep residual network architecture. 

\subsubsection{The functionality of the deep dense neural network}

The ablation experiment that demonstrates the role of the dense neural network in the DCFAE model is conducted by removing $DeepDenseNet_{\boldsymbol{\xi}}$ and then directly performing k-means clustering on the $\boldsymbol{\mu}$ vectors outputted by the encoder. The truncated model is marked as DCFAE - DenNet and the clustering results are manifested in Table \ref{Tb-498-5}.

\begin{table*}[!h]
\centering
\fontsize{7.2}{13}\selectfont
\setlength{\tabcolsep}{5.7pt}
\setlength{\arrayrulewidth}{0.3mm}
\caption{\fontsize{10pt}{\baselineskip}\selectfont Clustering results w.r.t. DCFAE - DenNet and DCFAE}
\begin{tabular}{l ccc ccc ccc ccc ccc}
\toprule
 & \multicolumn{3}{c}{MNIST} & \multicolumn{3}{c}{MNIST-test} & \multicolumn{3}{c}{USPS} & \multicolumn{3}{c}{Fashion-MNIST} & \multicolumn{3}{c}{COIL-100}\\
\cmidrule(lr){2-4} \cmidrule(lr){5-7} \cmidrule(lr){8-10} \cmidrule(lr){11-13} \cmidrule(lr){14-16}
Model & ACC & NMI & ARI & ACC & NMI & ARI & ACC & NMI & ARI & ACC & NMI & ARI & ACC & NMI & ARI \\
\midrule
DCFAE - DenNet & 85.98 & 76.06 & 73.80 & 85.15 & 75.41 & 72.25 & 79.15 & 79.26 & 73.99 & 61.53 & 57.26 & 45.68 & 52.67 & 77.13 & 45.64 \\
DCFAE & \textbf{97.36} & \textbf{93.29} & \textbf{94.26} & \textbf{97.56} & \textbf{93.70} & \textbf{94.70} & \textbf{97.30} & \textbf{93.03} & \textbf{94.72} & \textbf{64.34} & \textbf{68.26} & \textbf{52.40} & \textbf{82.79} & \textbf{91.39} & \textbf{77.10} \\
\bottomrule
\end{tabular}
\label{Tb-498-5}
\end{table*}

\begin{figure*}[!ht]
\centering
\subfigure[Latent space of DCFAE-DenNet on MNIST-test]{
\begin{minipage}[t]{0.31\textwidth}
\includegraphics[height=4.6cm,width=5.2cm]{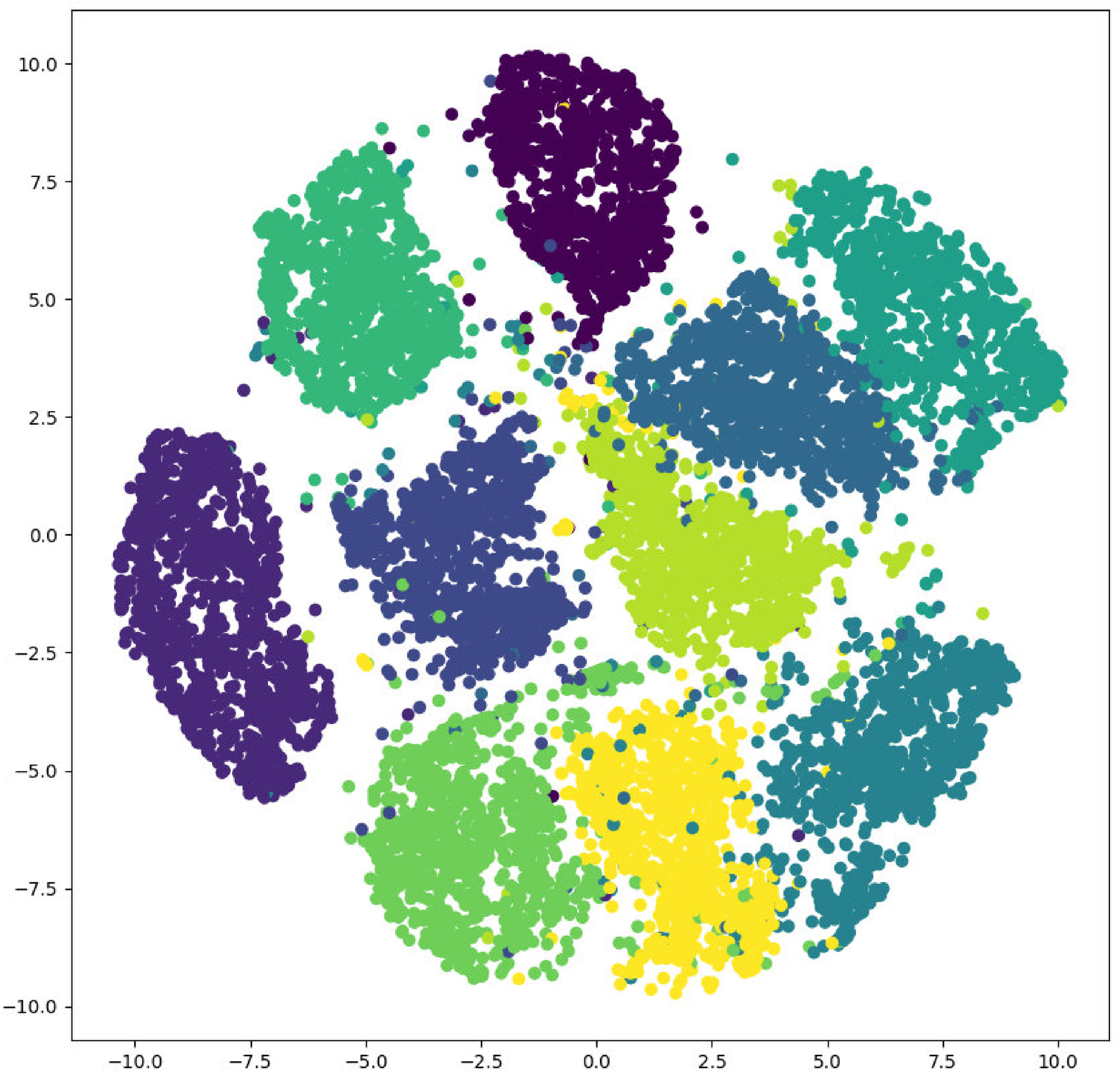}
\end{minipage}
}
\subfigure[Latent space of DCFAE-DenNet on USPS]{
\begin{minipage}[t]{0.31\linewidth}
\includegraphics[height=4.6cm,width=5.2cm]{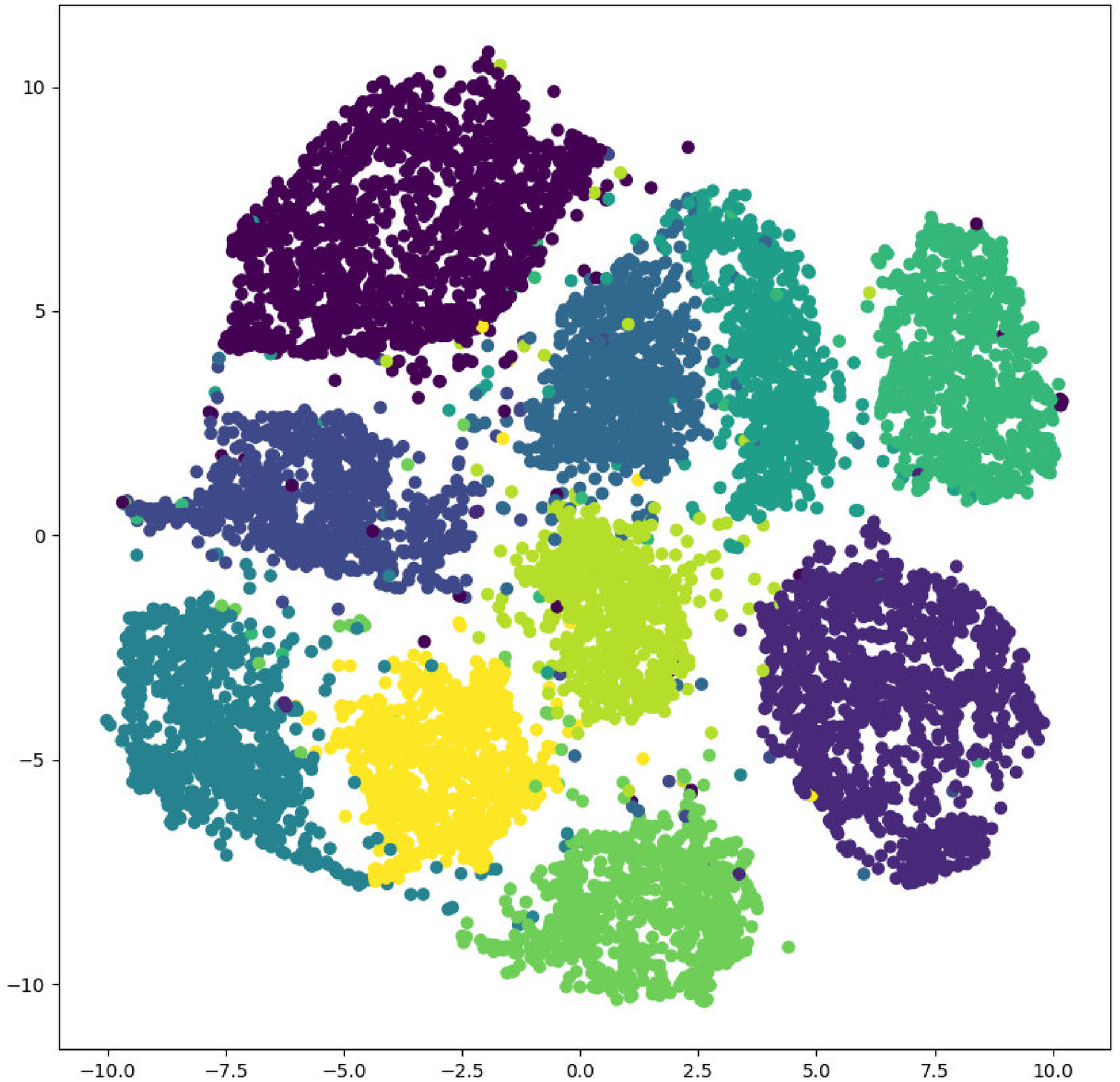}
\end{minipage}
}
\subfigure[Latent space of DCFAE-DenNet on COIL-100]{
\begin{minipage}[t]{0.31\linewidth}
\includegraphics[height=4.6cm,width=5.2cm]{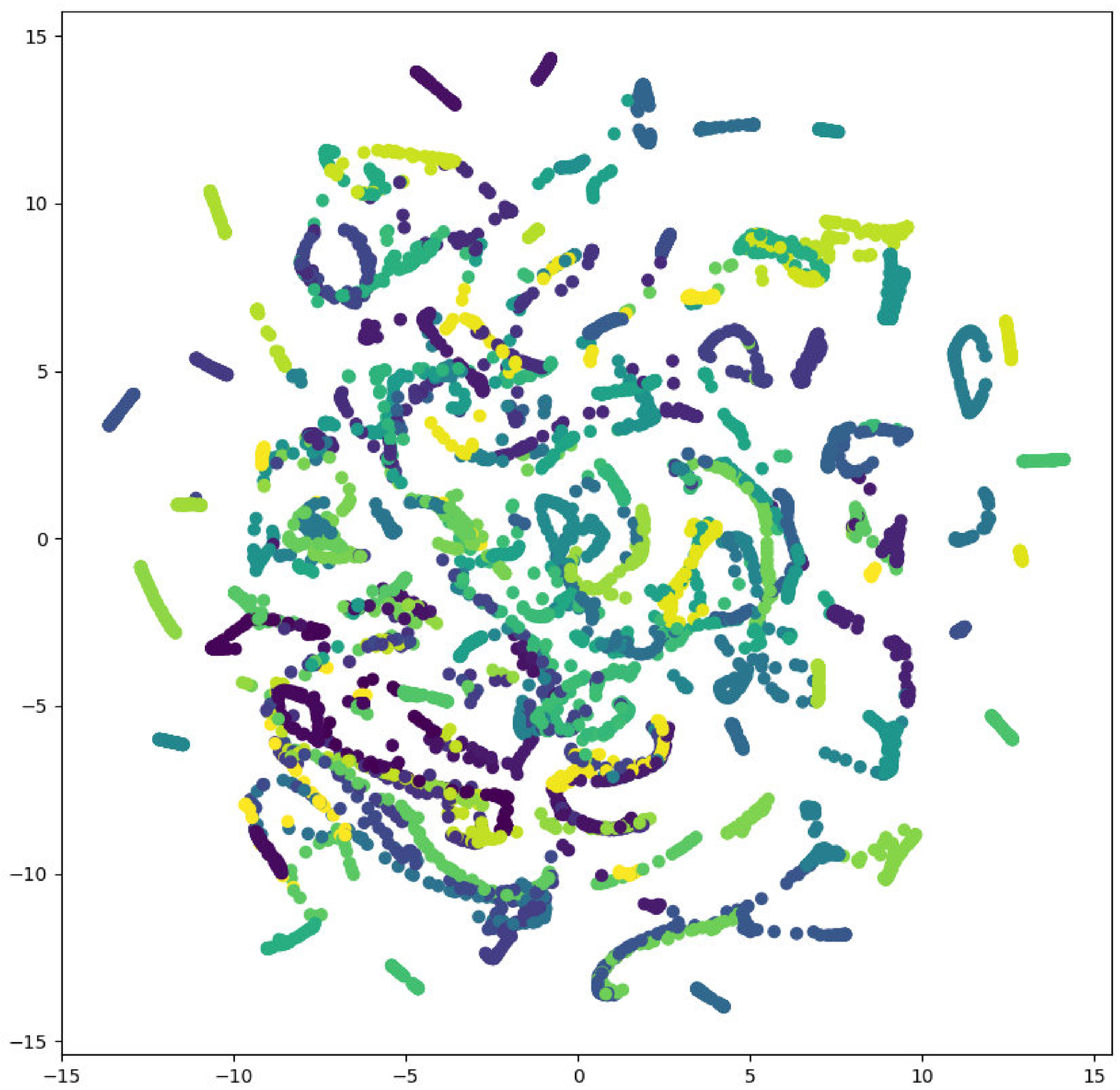}
\end{minipage}
}

\subfigure[Latent space of DCFAE on MNIST-test]{
\begin{minipage}[t]{0.31\linewidth}
\includegraphics[height=4.6cm,width=5.2cm]{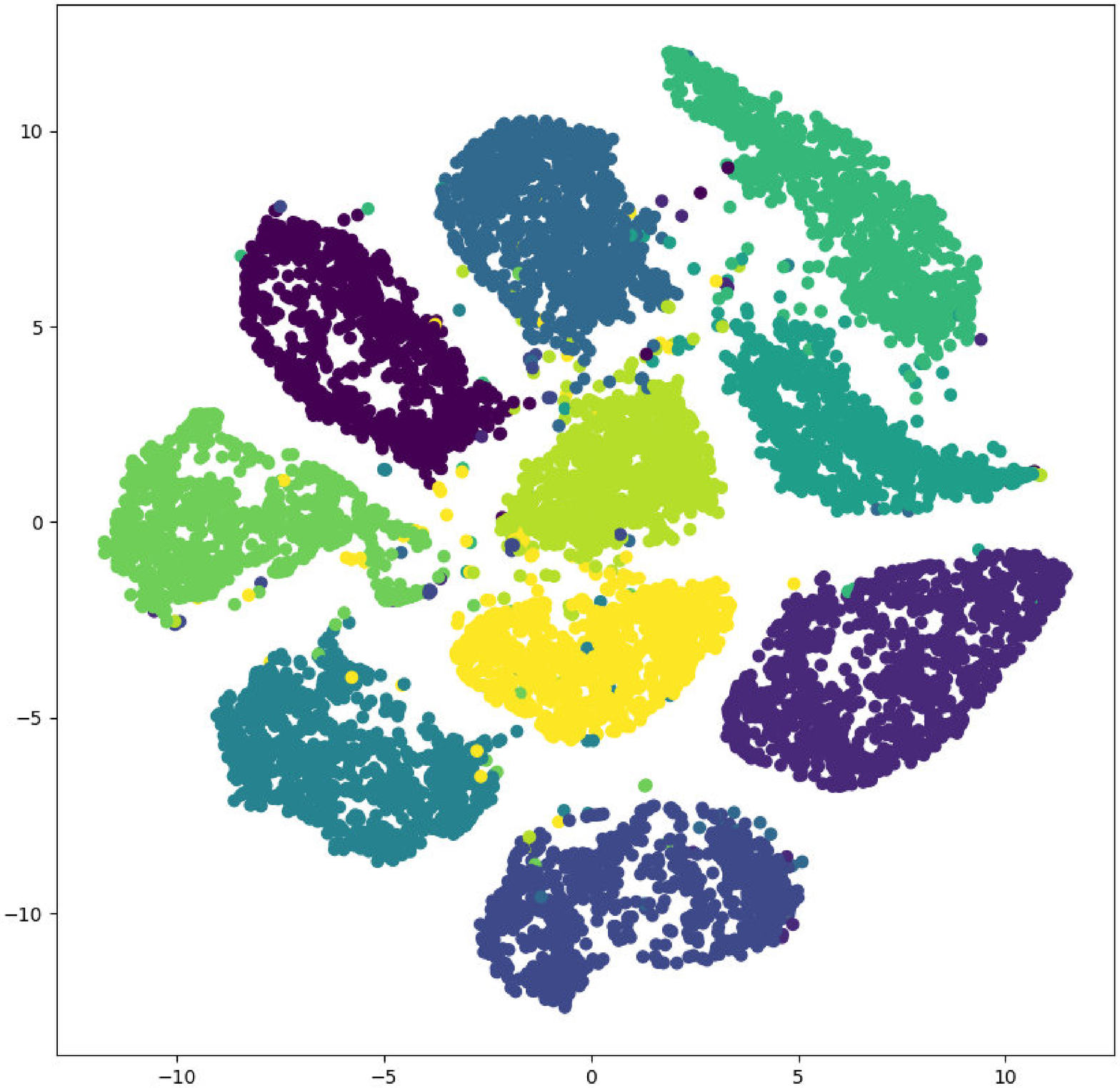}
\end{minipage}
}
\subfigure[Latent space of DCFAE on USPS]{
\begin{minipage}[t]{0.31\linewidth}
\includegraphics[height=4.6cm,width=5.2cm]{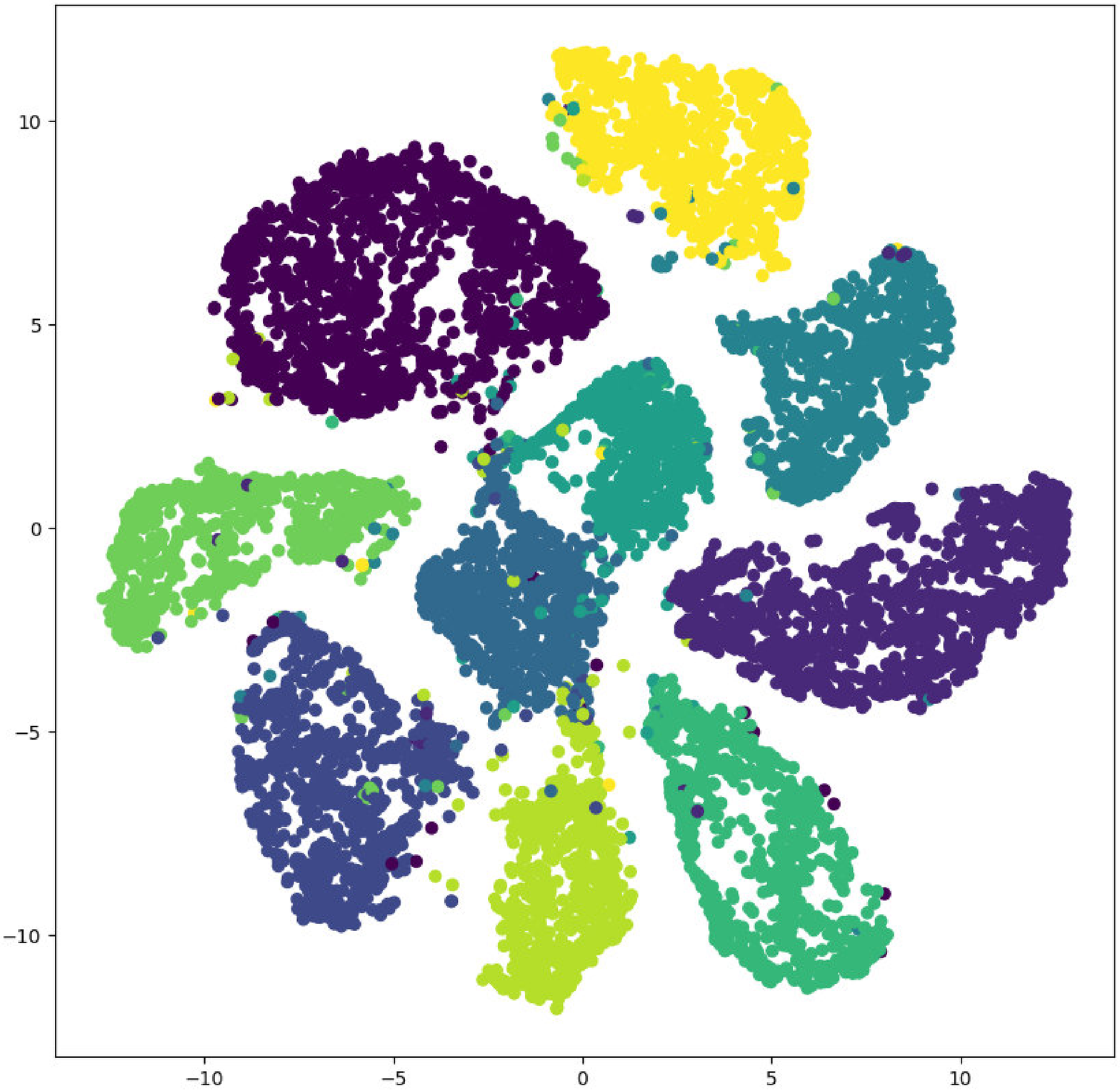}
\end{minipage}
}
\subfigure[Latent space of DCFAE on COIL-100]{
\begin{minipage}[t]{0.31\linewidth}
\includegraphics[height=4.6cm,width=5.2cm]{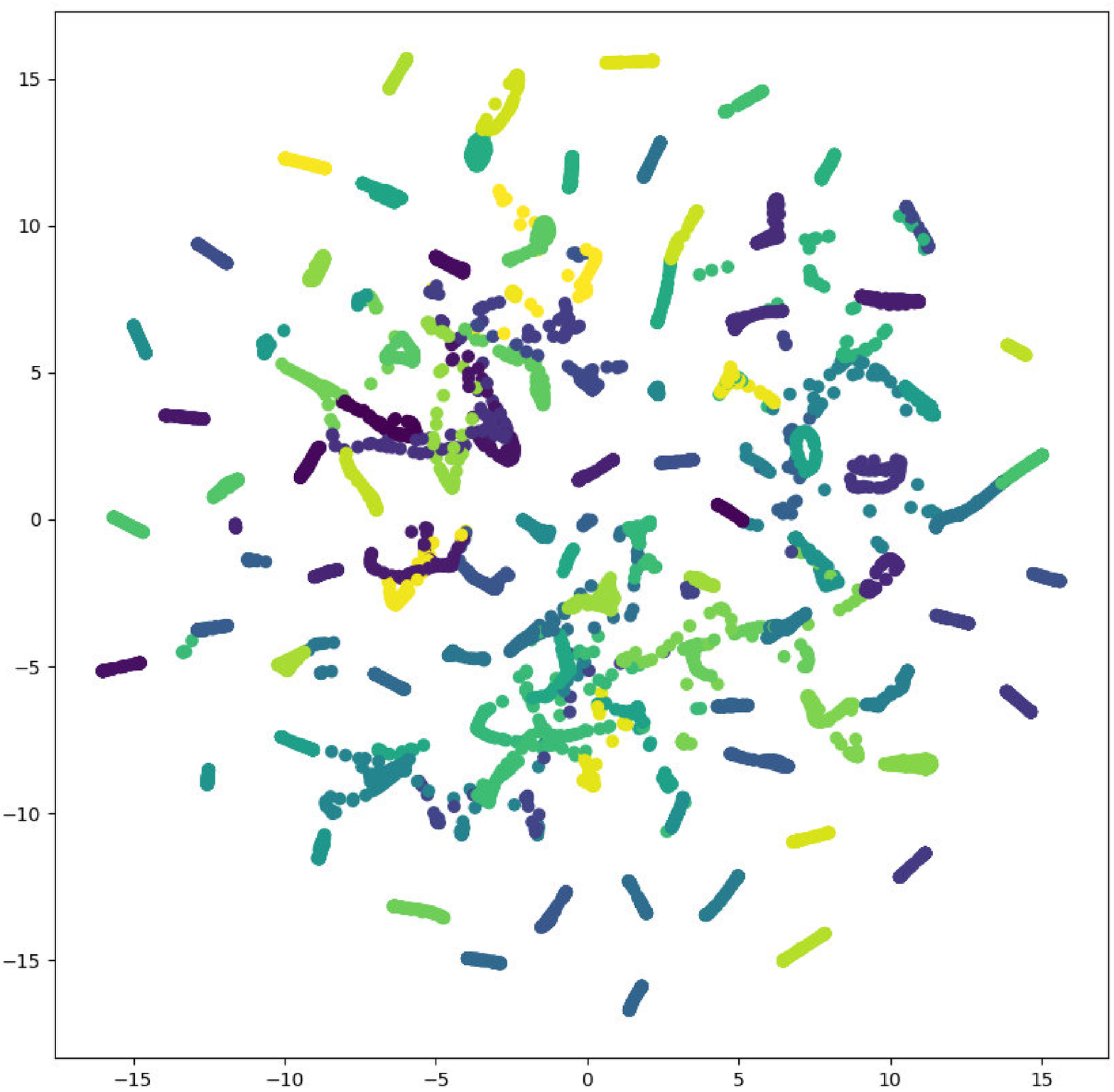}
\end{minipage}
}

\caption{Latent spaces w.r.t. the DCFAE - DenNet and DCFAE models. Contrasted with DCFAE - DenNet, individual clusters are more compact and the distances between different clusters are larger in the latent space of DCFAE.}
\label{Pic-546}
\end{figure*}

From the table, we can see that there are drastic clustering performance boosts when FAE is equipped with the $DeepDenseNet_{\boldsymbol{\xi}}$, suggesting that the deep dense neural network can truly learn a k-means clustering friendly embedding space which agglomerates intra-cluster data points and enlarges the distances among different clusters by minimizing the cross-entropy between the $P$ and $Q$ distributions. Besides, the $\boldsymbol{\mu}$ and $\boldsymbol{c}$ latent spaces of three datasets plotted by t-SNE in Fig. \ref{Pic-546} bolster the assertion.

\subsection{Monitoring the adversarial training process}

Training FAE could be difficult since it contains a discriminator that competes with the generator during the training process, which means the FAE may fail to converge (i.e., the discriminator and the generator cannot reach the equilibrium state) when one of the discriminator and generator dominates the other. Hence, we monitor the adversarial training process through two scores that point out how well the discriminator and generator perform as the training progressing. More specifically, the discriminator score is the probability that the genuineness of an image is correctly identified by the discriminator, which is defined by the following equation:

\begin{equation}\label{Eq-676-13}
Discriminator\ score = \frac{1}{2M} \sum\nolimits_{i=1}^M \left[S\left(Dis\left(\boldsymbol{x}^{(i)}\right)\right) + \left( 1 - S\left( Dis\left( \tilde{\boldsymbol{x}}^{(i)} \right) \right) \right)\right],
\end{equation}

\noindent where $M$ stands for the batch size, $S(x) = \frac{1}{1 + e^{-x}}$ is the sigmoid function, $Dis\left(\boldsymbol{x}^{(i)}\right)$ and $Dis\left( \tilde{\boldsymbol{x}}^{(i)} \right)$ denote the outputs of the discriminator for a real sample $\boldsymbol{x}^{(i)}$ and the corresponding $\tilde{\boldsymbol{x}}^{(i)}$ that reconstructed by the generator.

While the generator score tells the probability that the discriminator classifies a reconstructed image as a real one, which is expressed below:

\begin{equation}\label{Eq-685-14}
Generator\ score = \frac{1}{M} \sum\nolimits_{i=1}^M S\left( Dis\left( \tilde{\boldsymbol{x}}^{(i)} \right) \right).
\end{equation}

\begin{figure*}[!ht]
\centering
\subfigure[COIL-100]{
\begin{minipage}[t]{0.45\textwidth}
\includegraphics[height=5.5cm,width=7.315cm]{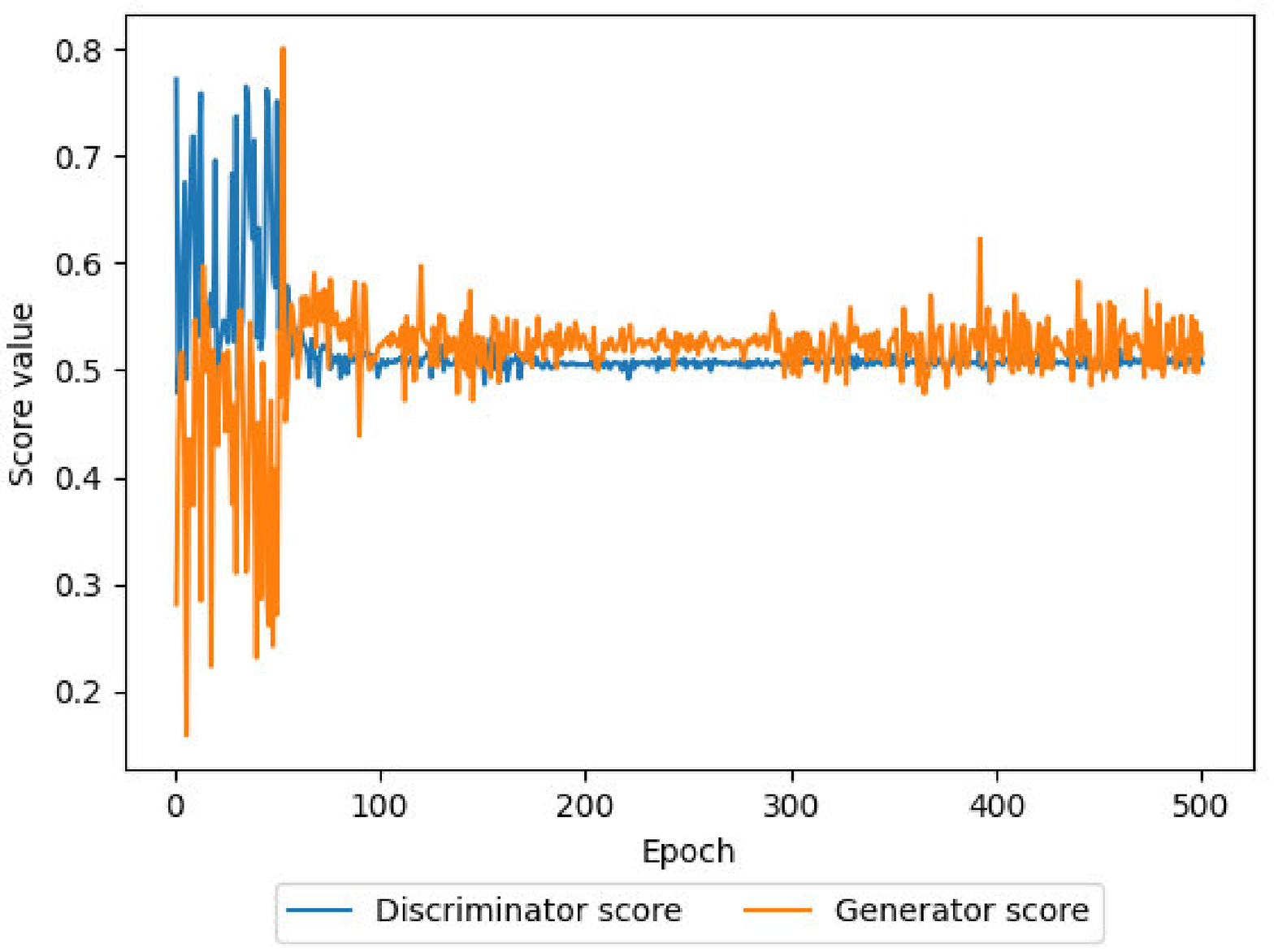}
\end{minipage}
}
\subfigure[USPS]{
\begin{minipage}[t]{0.45\linewidth}
\includegraphics[height=5.5cm,width=7.315cm]{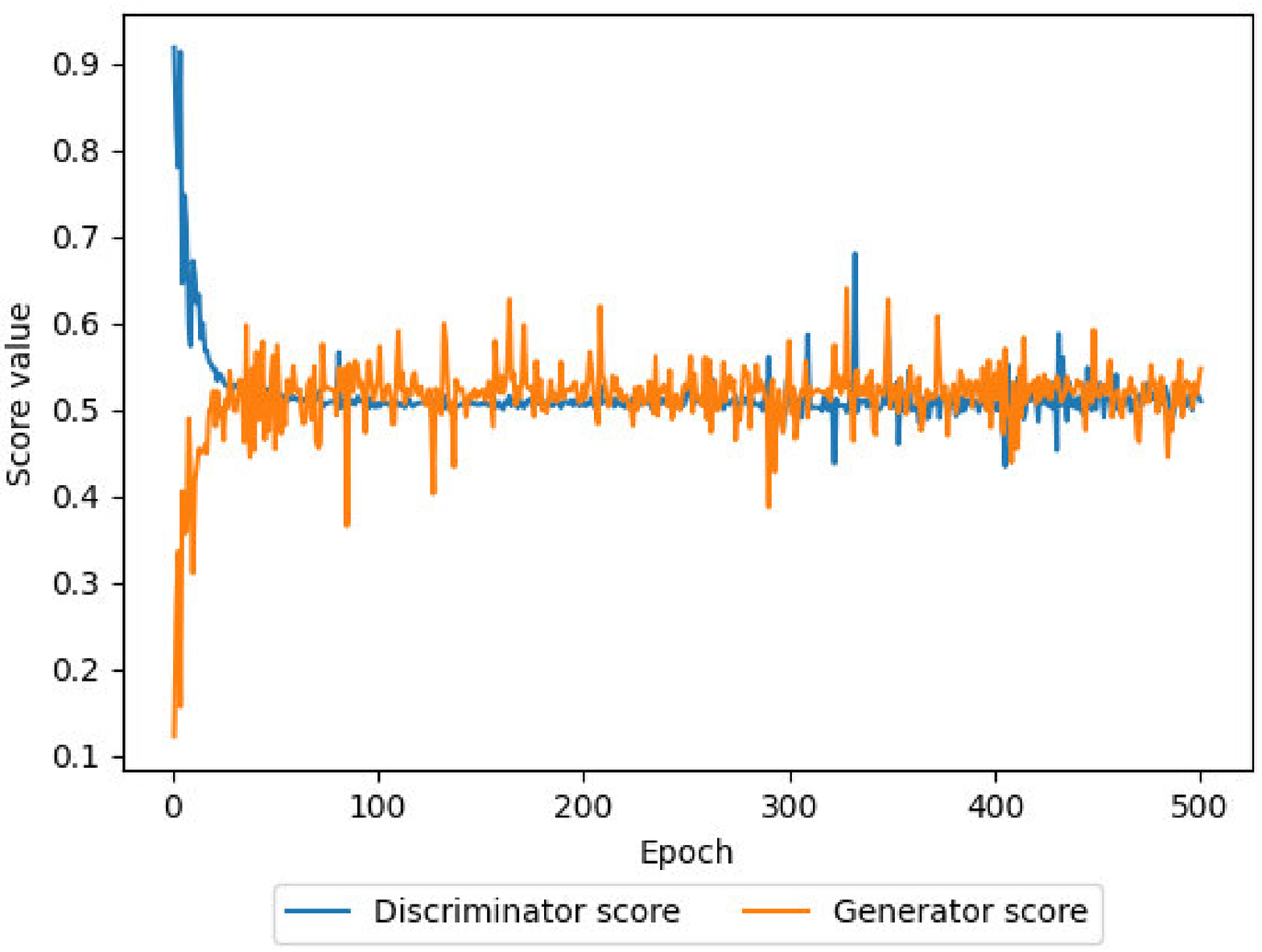}
\end{minipage}
}

\subfigure[MNIST-test]{
\begin{minipage}[t]{0.45\textwidth}
\includegraphics[height=5.5cm,width=7.315cm]{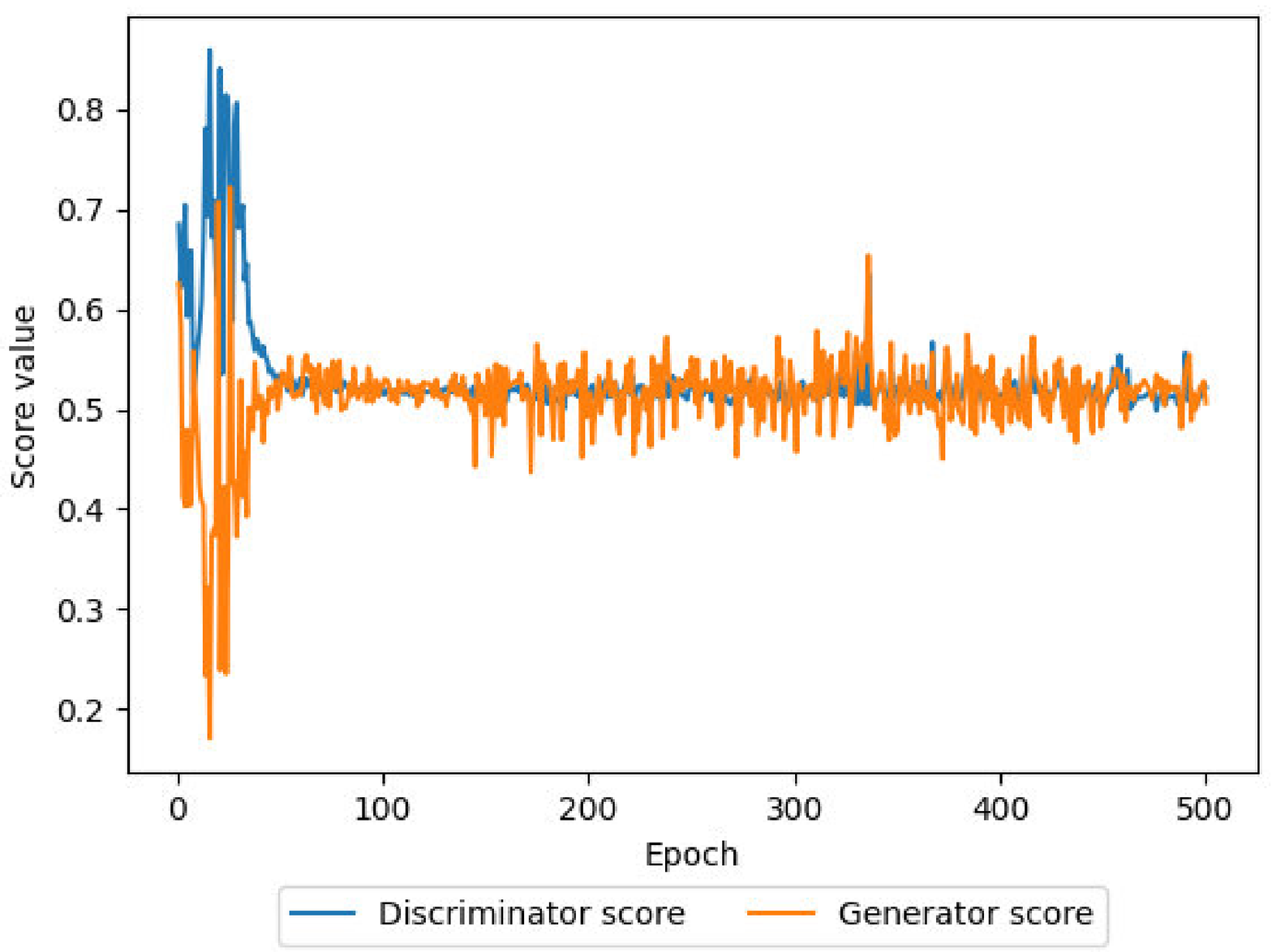}
\end{minipage}
}
\subfigure[MNIST]{
\begin{minipage}[t]{0.45\linewidth}
\includegraphics[height=5.5cm,width=7.315cm]{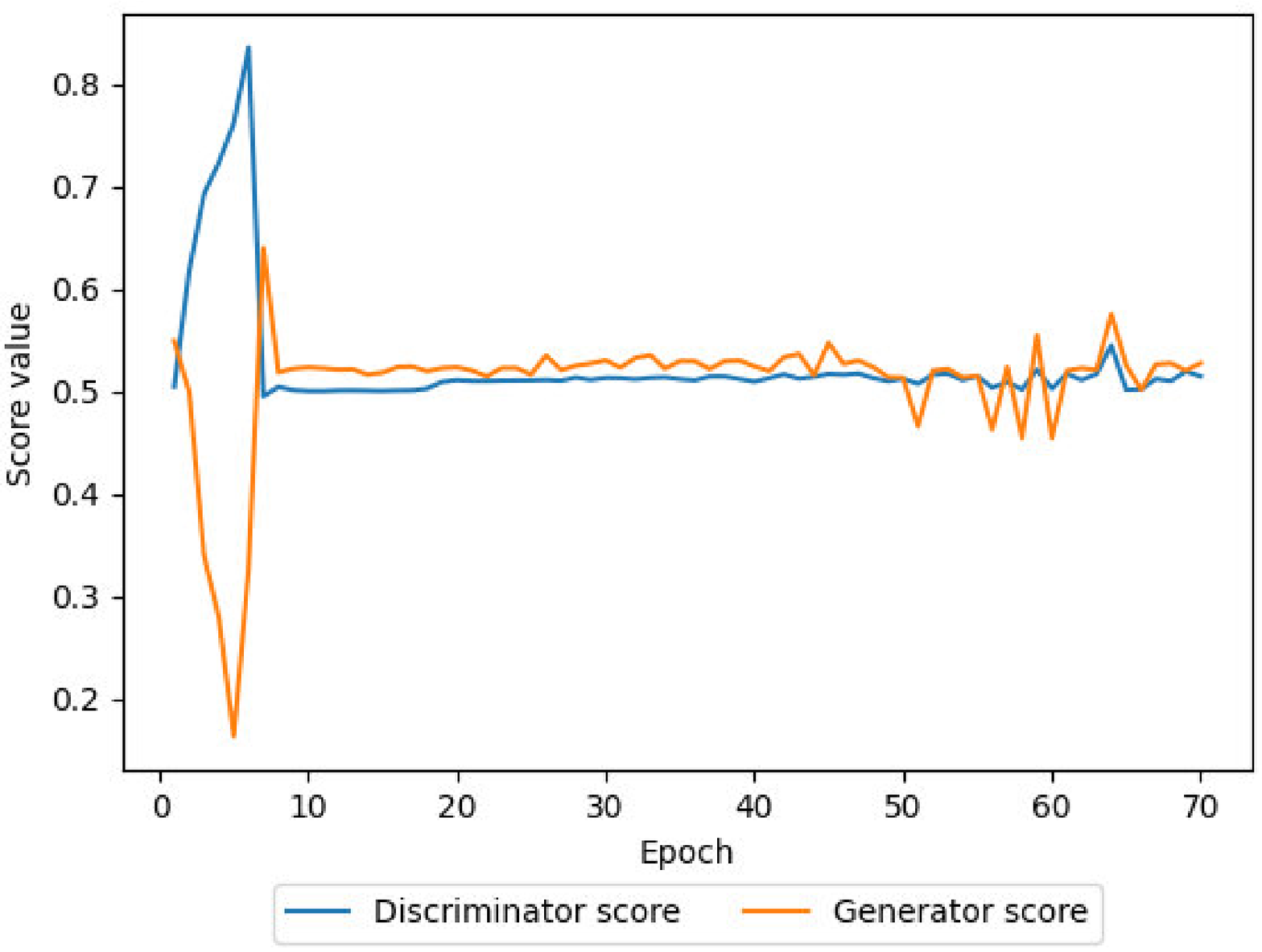}
\end{minipage}
}

\caption{Trending of the discriminator score and generator score on 4 datasets. (a) COIL-100, (b) USPS, (c) MNIST-test, (d) MNIST.}
\label{Pic-731}
\end{figure*}

Under the ideal condition, both the discriminator score and the generator score would be 0.5 when reaching the equilibrium, since the discriminator cannot figure out the difference between real and fake images. To investigate the adversarial behavior of the designed FAE framework during the training, we train FAE on 4 datasets and the line graphs of the two scores are plotted in Fig. \ref{Pic-731}.

Fig. \ref{Pic-731} shows that there is an intensive oscillation among the two scores at the beginning of the training process, implying that both the discriminator and the generator networks change rapidly as they try to compete with each other. But after about 50 epochs, all curves start to stabilize at 0.5 around, which means the discriminator and generator have reached the equilibrium state. These figures prove that the designed FAE architecture could successfully converge under the adversarial training process.
\subsection{Parameter analysis} 

DCFAE generally contains four parameters: balance parameters $\lambda$ and ${\lambda}'$, the number of degrees of freedom $\rho$ in Eq. \eqref{Eq-274-9}, and the dimension of the latent variable $L$. Impacts of these parameters to the clustering performance are examined through the following experiments.

\subsubsection{Balance parameters}

\begin{figure*}[!ht]
\centering
\subfigure[MNIST]{
\begin{minipage}[t]{0.45\textwidth}
\includegraphics[height=6cm,width=6.6cm]{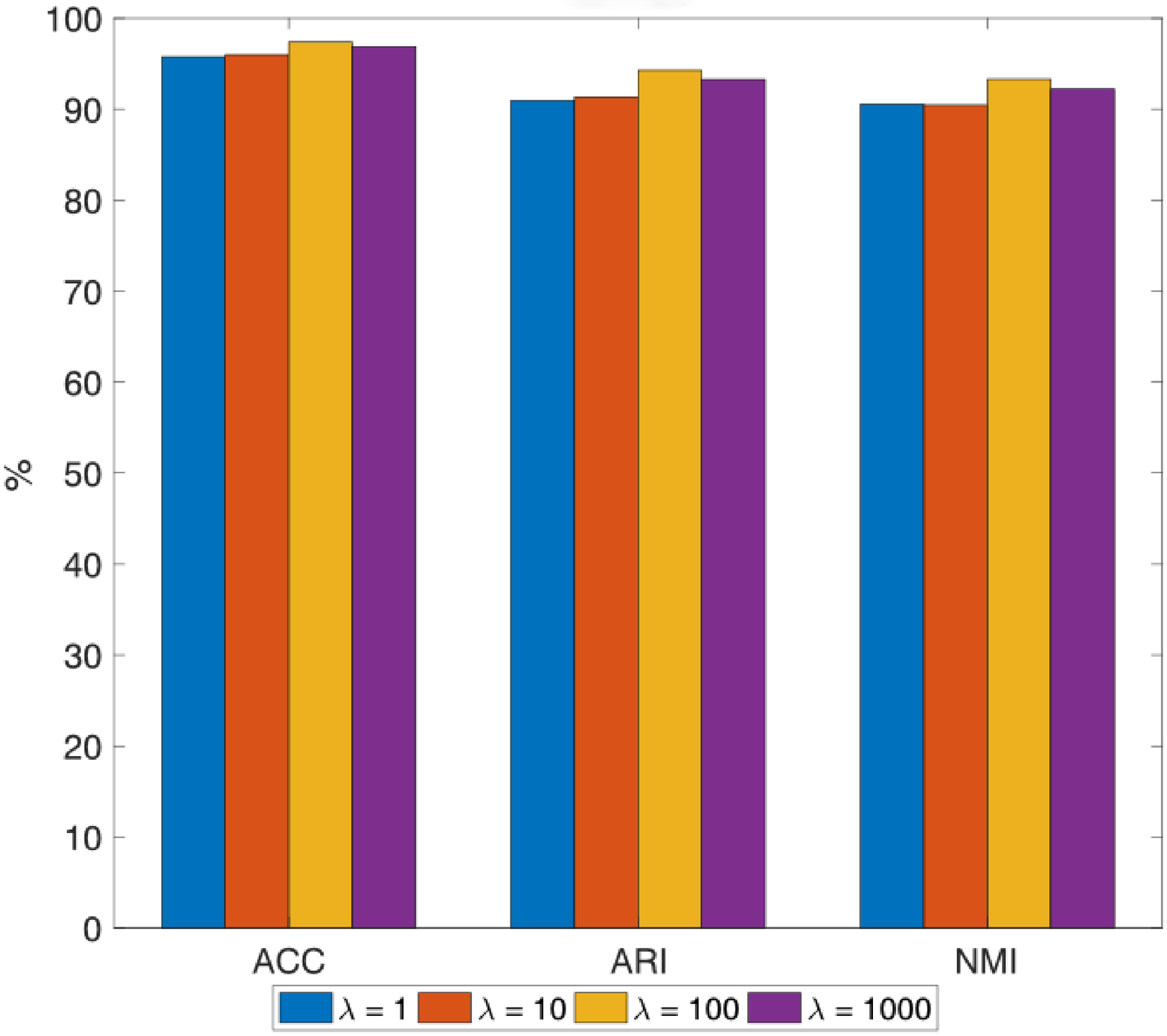}
\end{minipage}
}
\subfigure[MNIST-test]{
\begin{minipage}[t]{0.45\linewidth}
\includegraphics[height=6cm,width=6.6cm]{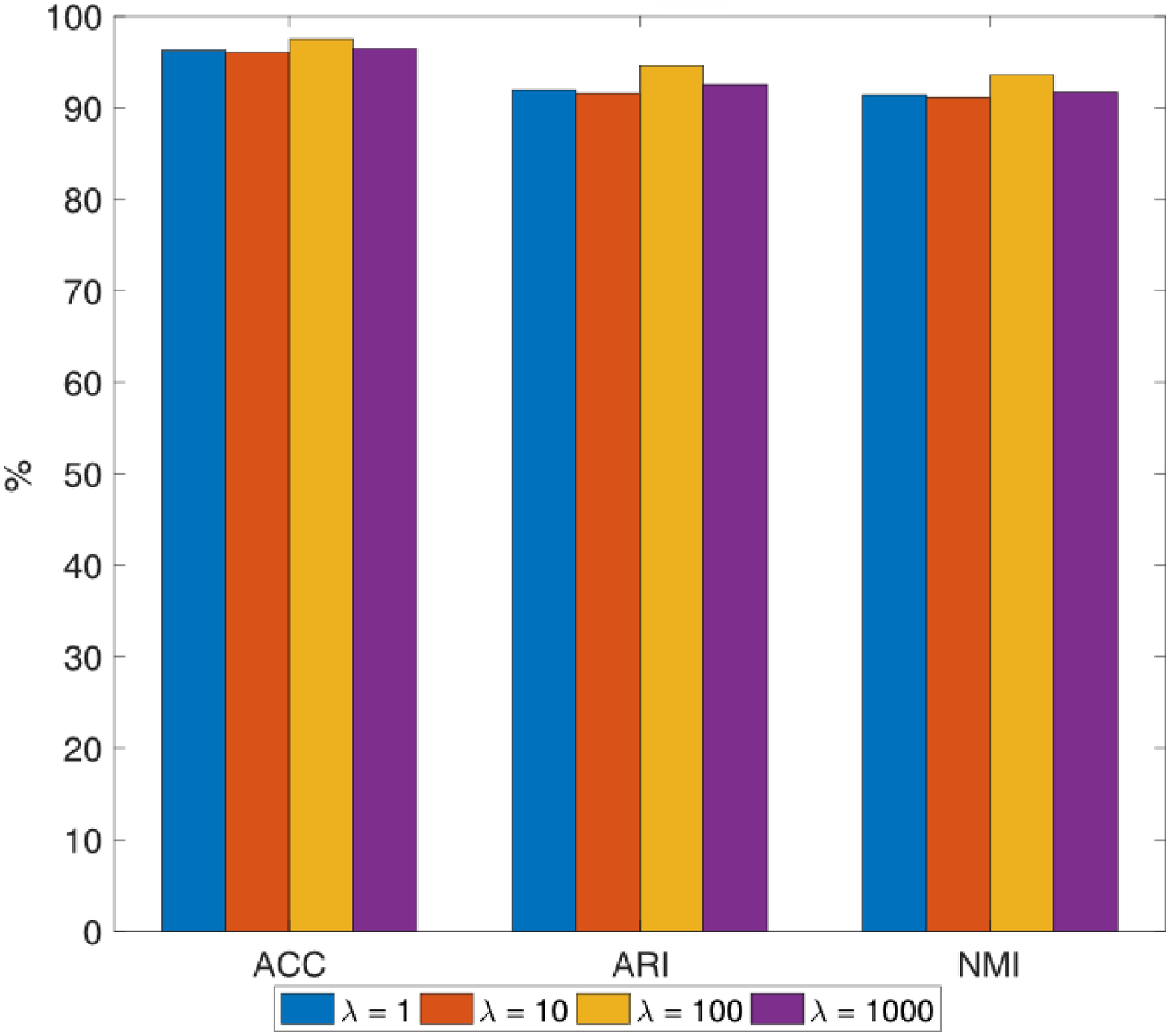}
\end{minipage}
}

\caption{Clustering performance w.r.t. the balance parameter $\lambda$ on two datasets. (a) MNIST, (b) MNIST-test.}
\label{Pic-733}
\end{figure*}

We first analyze the influence of $\lambda$ on the performance of DCFAE. The experiment is carried out by setting the value of $\lambda$ to 1, 10, 100, 1000 separately and fixing all other parameters to their default values stated in the 4.1 subsection. Clustering results on MNIST and MNIST-test datasets are shown in Fig. \ref{Pic-733}. It can be clearly observed from the figure that when $\lambda$ equals 100, DCFAE would gain the best clustering performance, suggesting that setting $\lambda$ to 100 could properly balance the adversarial loss and the VAE loss.

\begin{figure*}[!ht]
\centering
\subfigure[MNIST]{
\begin{minipage}[t]{0.45\textwidth}
\includegraphics[height=6cm,width=6.6cm]{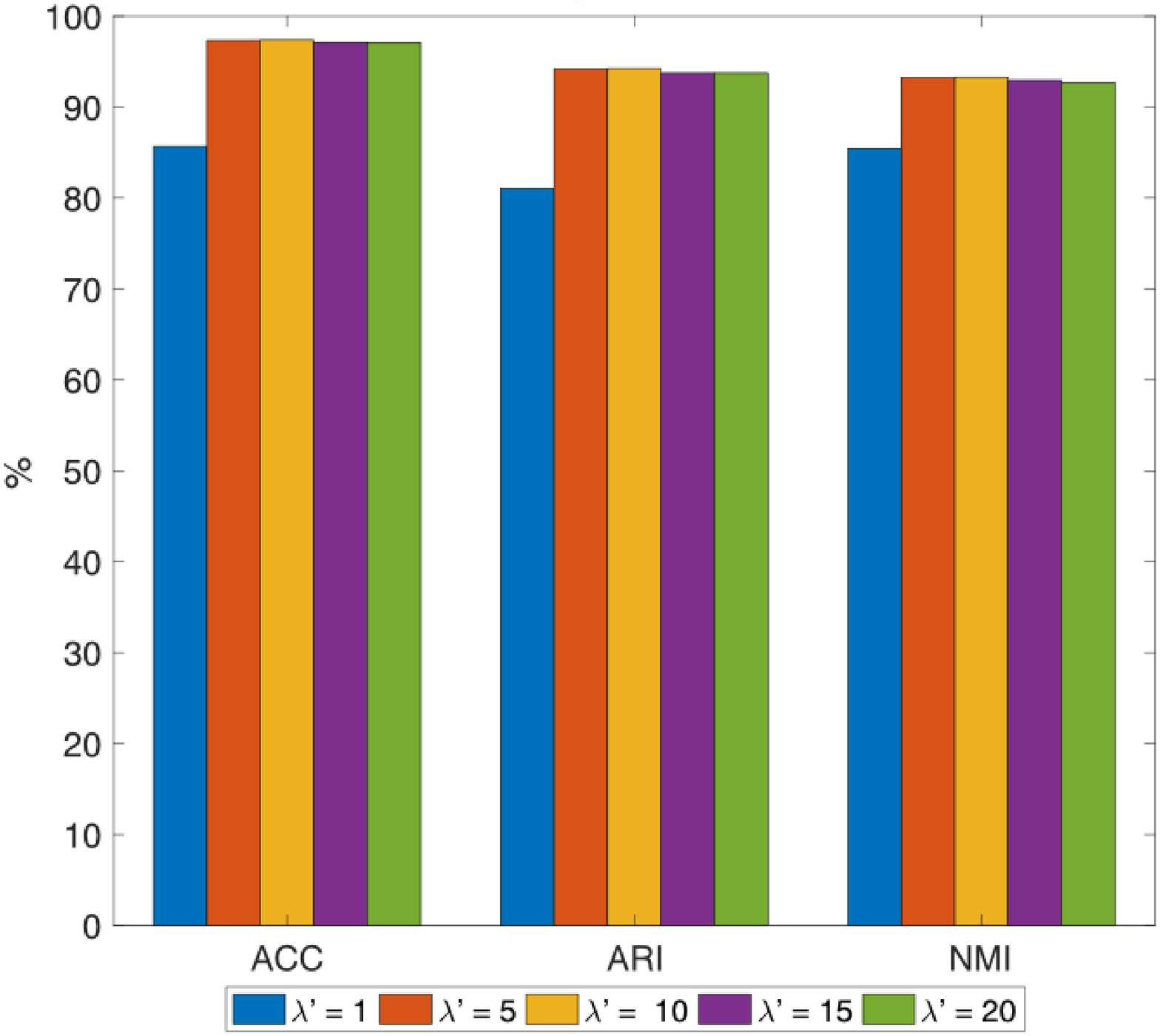}
\end{minipage}
}
\subfigure[USPS]{
\begin{minipage}[t]{0.45\linewidth}
\includegraphics[height=6cm,width=6.6cm]{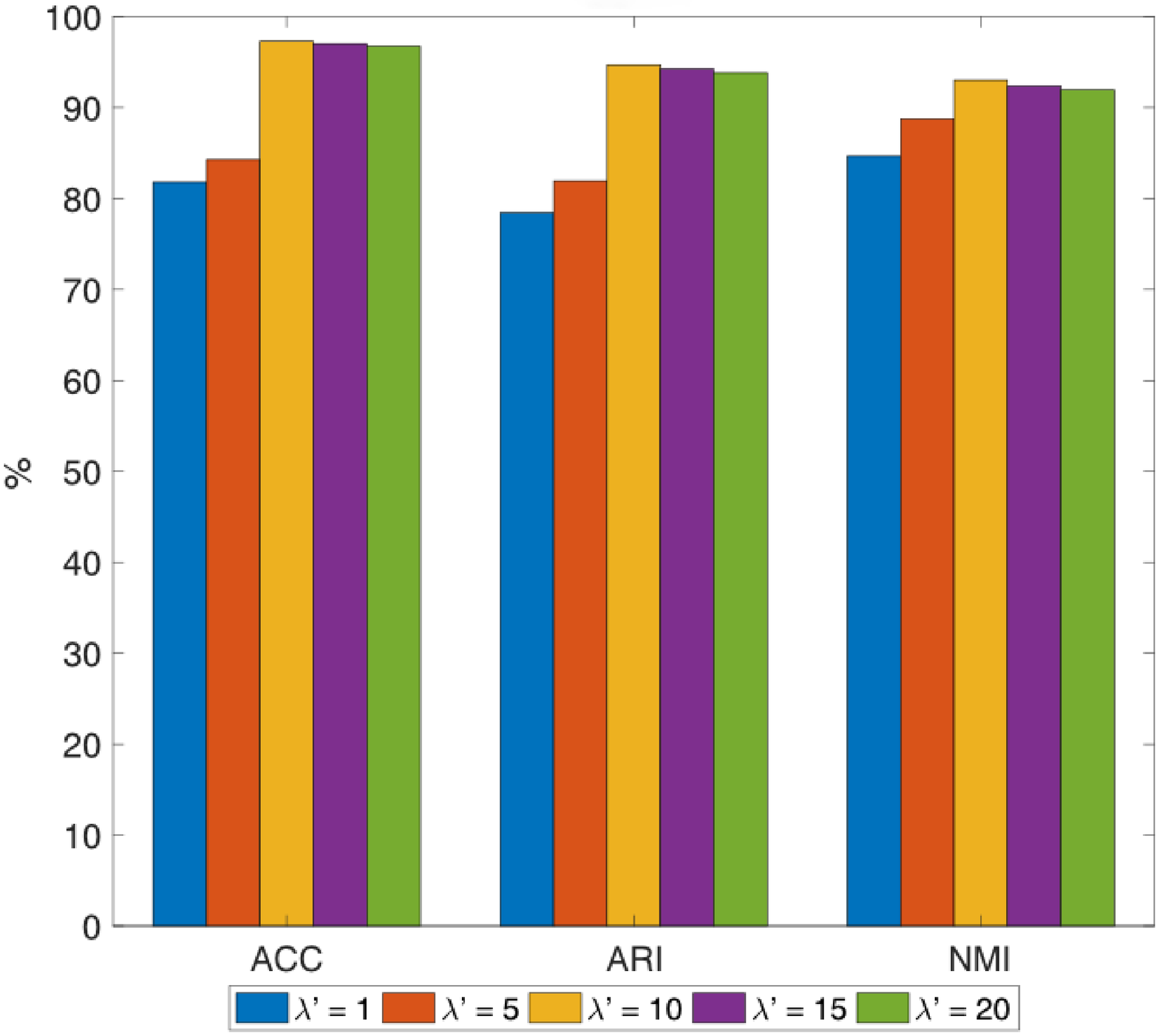}
\end{minipage}
}

\caption{Clustering performance w.r.t. the balance parameter $\lambda'$ on two datasets. (a) MNIST, (b) USPS.}
\label{Pic-752}
\end{figure*}

The experiment that unveils the impact of $\lambda'$ on the performance of DCFAE is akin to the former one, except that the value of $\lambda'$ varies in 1, 5, 10, 15, and 20. Fig. \ref{Pic-752} illustrates the evaluation results of DCFAE tested on the MNIST and USPS datasets. The figure indicates that assigning 1 to $\lambda'$ would lead to poor clustering performance, which means $\mathcal{L}_\gamma$ has a weak effect in Eq. \eqref{Eq-294-12a}, causing the deep dense neural network cannot be sufficiently optimized. While on the other hand, the clustering performance would slightly shrink when $\lambda'$ exceeds 10. Based on these clues, we choose 10 as the optimal value for $\lambda'$.

\subsubsection{The number of degrees of freedom}

\begin{figure*}[!ht]
\centering
\subfigure[MNIST]{
\begin{minipage}[t]{0.45\textwidth}
\includegraphics[height=6cm,width=6.6cm]{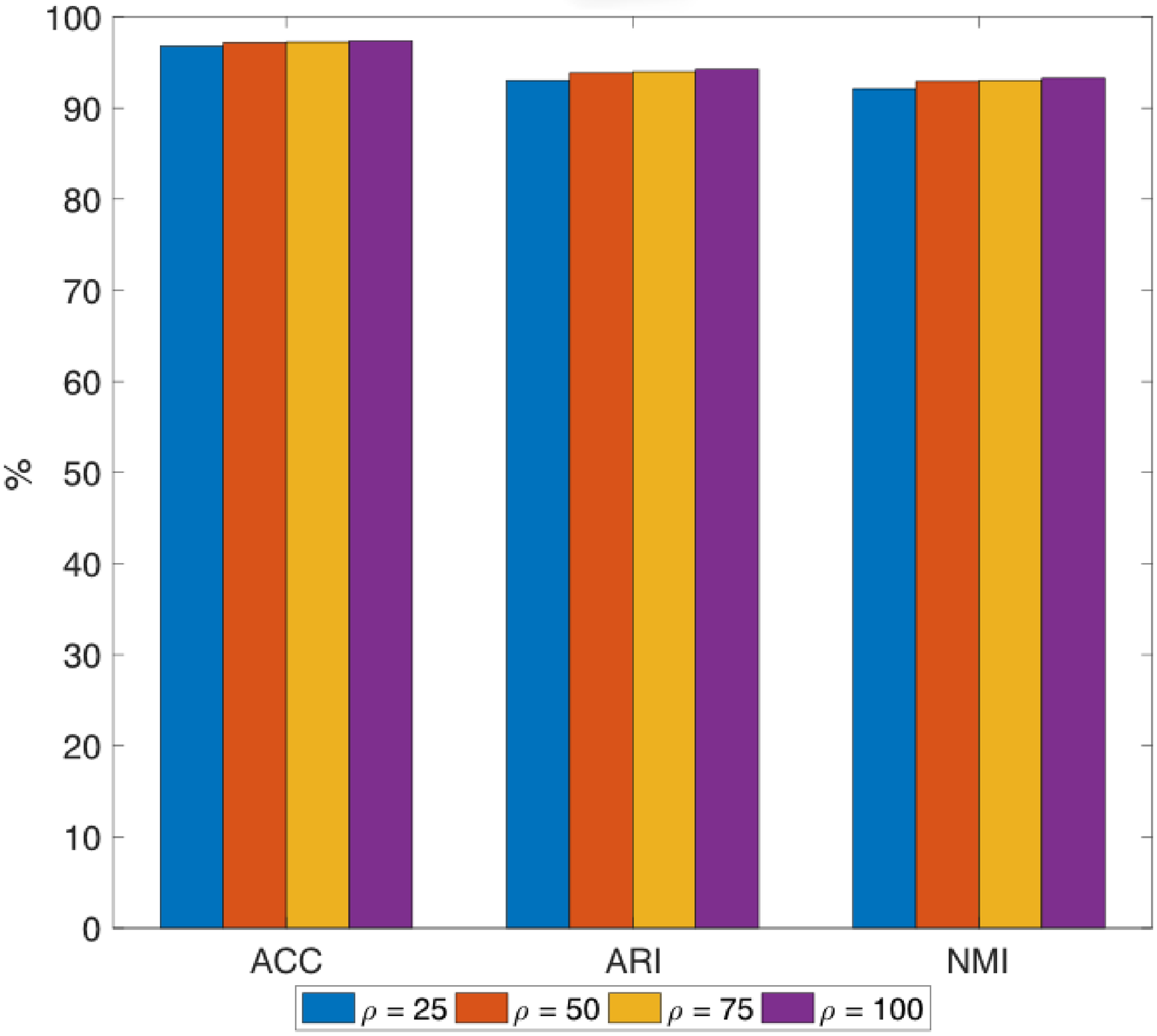}
\end{minipage}
}
\subfigure[USPS]{
\begin{minipage}[t]{0.45\linewidth}
\includegraphics[height=6cm,width=6.6cm]{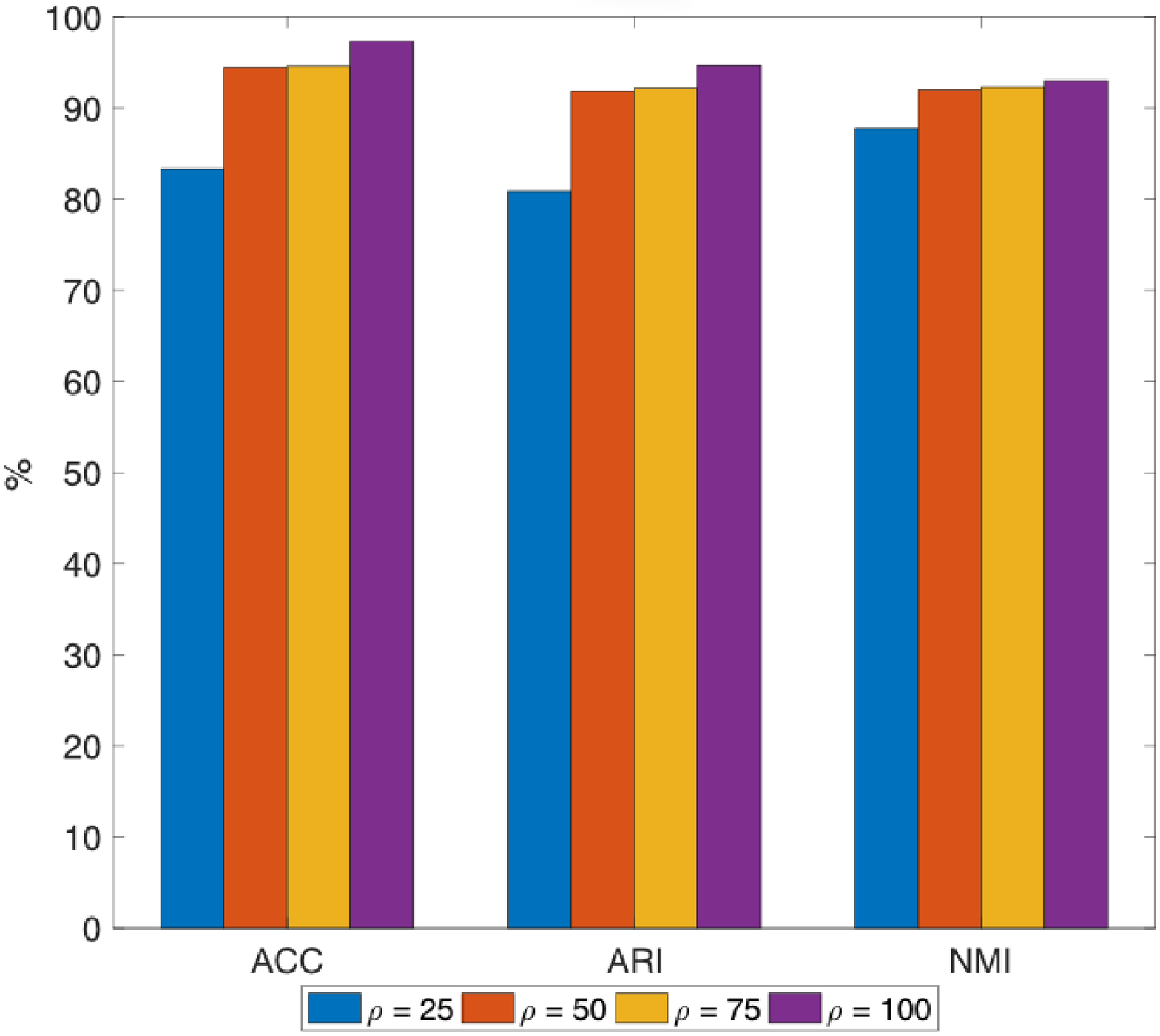}
\end{minipage}
}

\caption{Clustering performance w.r.t. the number of degrees of freedom $\rho$ on two datasets. (a) MNIST, (b) USPS.}
\label{Pic-772}
\end{figure*}

As \cite{pmlr-v5-maaten09a} suggested, $\rho$ should have a relatively large value to capture the local structure of the data inside the latent space. Thus we change the value of $\rho$ from 25 to 100 in steps of 25 and set all other parameters as optimal to monitor the clustering performance. Experiment results on the MNIST and USPS datasets are demonstrated in Fig. \ref{Pic-772}. The figure displays that when the value of $\rho$ grows, clustering results would raise as consequence, manifesting the claim that large $\rho$ could better perceive the local structure of the latent space therefore benefiting the clustering process.

\subsubsection{The dimension of the latent variable}

\begin{figure*}[!ht]
\centering
\subfigure[MNIST]{
\begin{minipage}[t]{0.45\textwidth}
\includegraphics[height=6cm,width=6.6cm]{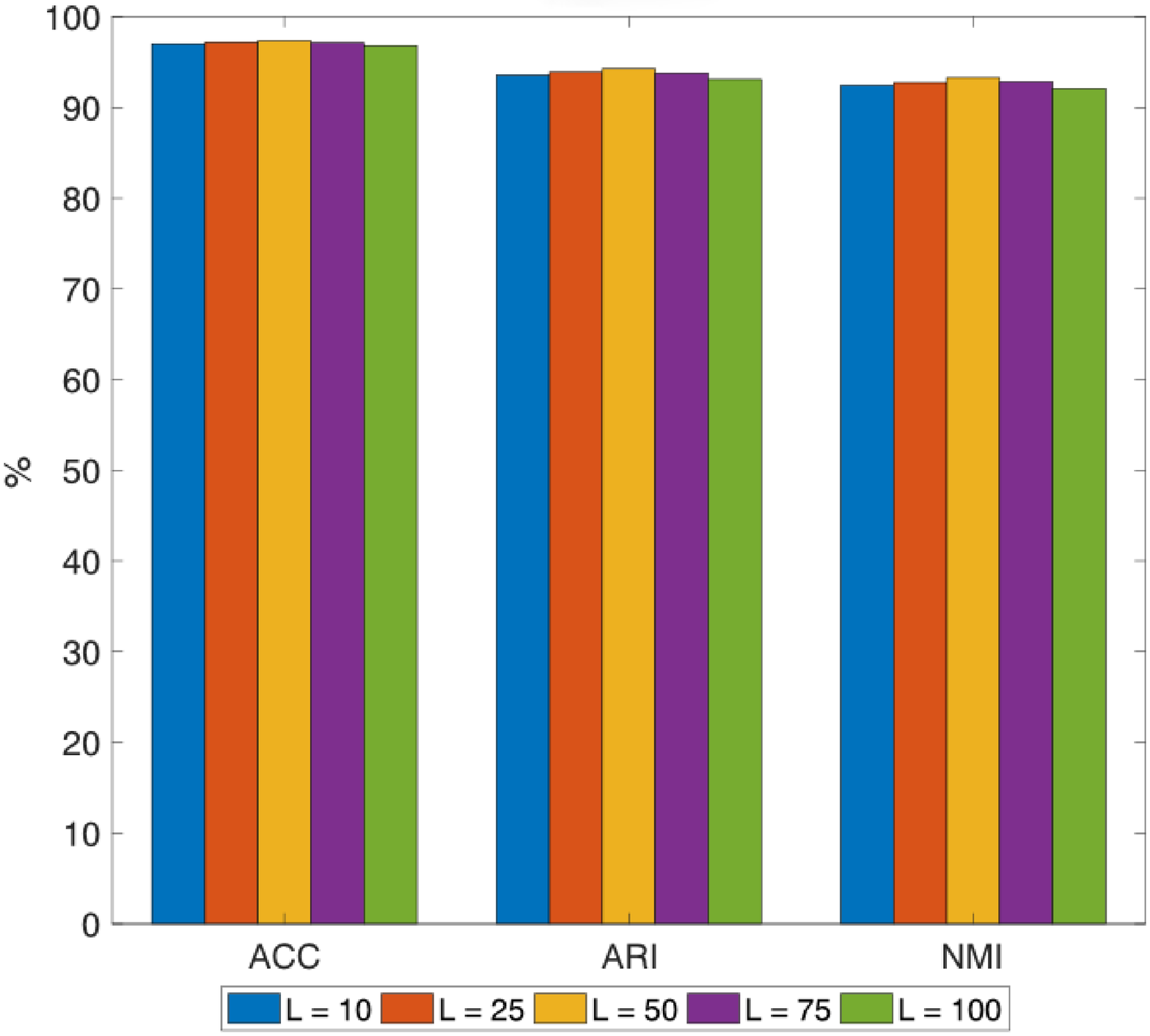}
\end{minipage}
}
\subfigure[USPS]{
\begin{minipage}[t]{0.45\linewidth}
\includegraphics[height=6cm,width=6.6cm]{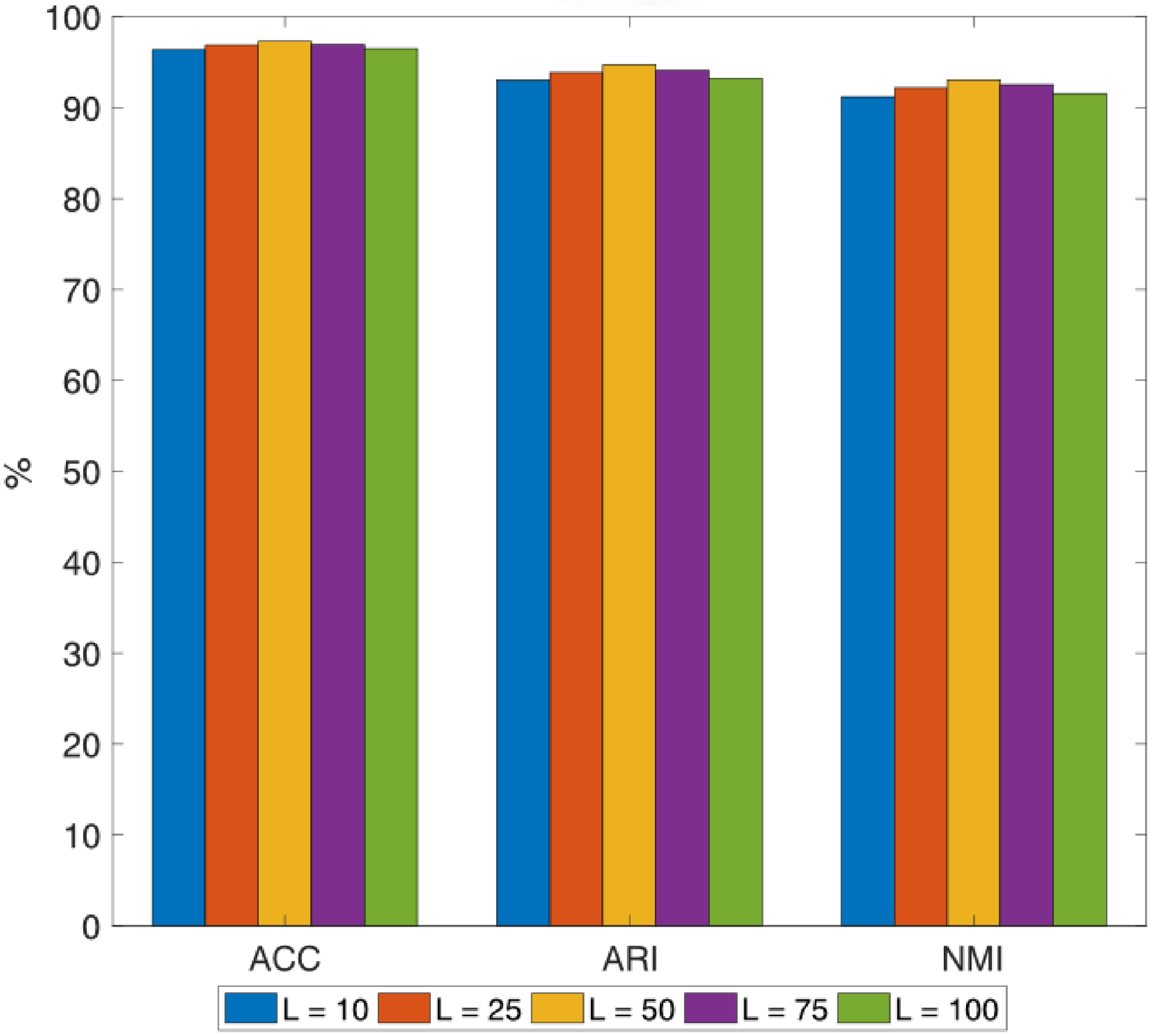}
\end{minipage}
}

\caption{Clustering performance w.r.t. the dimension of the latent variable $L$ on two datasets. (a) MNIST, (b) USPS.}
\label{Pic-795}
\end{figure*}

Finally, we inspect the sensitivity of DCFAE to the dimension of the latent variable through sequentially configuring $L$ to 10, 25, 50, 75, 100 and leaving other parameters unchanged. Experiment results on the MNIST and USPS datasets are expressed in Fig. \ref{Pic-795}. From the figure, we find out that $L$ has a moderate impact on the clustering performance and neither large nor small value of $L$ would yield better results. Hence we pick 50 for the default value of $L$. 

\section{Conclusion}

In this paper, a new DC method abbreviated as DCFAE is proposed by first amalgamating VAE with GAN to build the devised FAE framework and then making use of the deep dense neural network for learning an embedding space suitable for k-means clustering. Specifically, the VAE is coupled with GAN to improve the representation learning capability and procure more discriminative information that benefits the clustering operation. In addition, the FAE is fabricated with the deep residual network which enables DCFAE to catch higher-level features and therefore increases the clustering performance. Finally, inter-clusters and intra-cluster distances of the embedding space learned by a deep dense neural network are magnified and reduced respectively through minimizing the cross-entropy between two distributions that measuring pairwise similarities among the latent data points. Experiments performed on several image datasets witness the contributions of these maneuvers to the clustering performance and demonstrate the effectiveness of DCFEA compared with other state-of-the-art DC models. One future work can be trying to replace the prior distribution of the FAE's encoder, so that the clustering results could be directly obtained from the latent variables.

\section*{Acknowledgements}

This work is completed when the author is studying for his master's degree in Southwest Jiaotong University. The author would like to thank the school for providing the hardware resources and anyone who offered constructive suggestions.

\bibliographystyle{elsarticle-num-names} 
\bibliography{Bib}

\end{document}